Yiye Zou (邹艺业) ; Tianyu Li (李天宇) ; Lin Lu (陆林); Jingyu Wang (王靖宇) ✉ ; Shufan Zou (邹舒帆); Laiping Zhang (张来平) ; Xiaogang Deng (邓小刚)

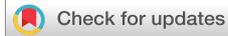

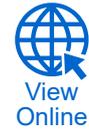
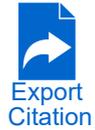

View Online  Export Citation

## Articles You May Be Interested In

Predicting unsteady incompressible fluid dynamics with finite volume informed neural network

*Physics of Fluids* (April 2024)

A high-order generalized differential quadrature method with lattice Boltzmann flux solver for simulating incompressible flows

*Physics of Fluids* (April 2023)

Physics-informed neural network based on a new adaptive gradient descent algorithm for solving partial differential equations of flow problems

*Physics of Fluids* (June 2023)

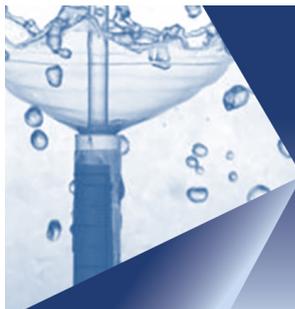

# Finite-difference-informed graph network for solving steady-state incompressible flows on block-structured grids

Yiye Zou (邹艺业),[1,2] Tianyu Li (李天宇),[1,2] Lin Lu (陆林),[1,2] Jingyu Wang (王靖宇),[2,3,a] Shufan Zou (邹舒帆),[4] Laiping Zhang (张来平),[2] and Xiaogang Deng (邓小刚)[5]

AFFILIATIONS

[1]School of Computer Science, Sichuan University, Chengdu 610065, China
[2]National Key Laboratory of Fundamental Algorithms and Models for Engineering Simulation, Sichuan University, Chengdu 610207, China
[3]School of Aeronautics and Astronautics, Sichuan University, Chengdu 610065, China
[4]College of Aerospace Science and Engineering, National University of Defense Technology, Changsha 410000, China
[5]Academy of Military Sciences, Beijing 100190, China

[a]Author to whom correspondence should be addressed: wangjingyu@scu.edu.cn

## ABSTRACT

Advances in deep learning have enabled physics-informed neural networks to solve partial differential equations. Numerical differentiation using the finite-difference (FD) method is efficient in physics-constrained designs, even in parameterized settings. In traditional computational fluid dynamics (CFD), body-fitted block-structured grids are often employed for complex flow cases when obtaining FD solutions. However, convolution operators in convolutional neural networks for FD are typically limited to single-block grids. To address this issue, graphs and graph networks are used to learn flow representations across multi-block-structured grids. A graph convolution-based FD method (GC-FDM) is proposed to train graph networks in a label-free physics-constrained manner, enabling differentiable FD operations on unstructured graph outputs. To demonstrate model performance from single- to multi-block-structured grids, the parameterized steady incompressible Navier–Stokes equations are solved for a lid-driven cavity flow and the flows around single and double circular cylinder configurations. When compared to a CFD solver under various boundary conditions, the proposed method achieves a relative error in velocity field predictions in the order of $10^{-3}$. Furthermore, the proposed method reduces training costs by approximately 20% compared to a physics-informed neural network. To further verify the effectiveness of GC-FDM in multi-block processing, a 30P30N airfoil geometry is considered, and the predicted results are reasonably compared with those given by CFD. Finally, the applicability of GC-FDM to a three-dimensional (3D) case is tested using a 3D cavity geometry.

## I. INTRODUCTION

Solving strongly nonlinear partial differential equations (PDEs) is an essential, but challenging, task in computational fluid dynamics due to the lack of closed-form solutions. Traditional numerical discretization methods have been developed to find approximate solutions. However, these methods often consume significant computational resources, especially when solving systems like the Navier–Stokes equations. In recent years, the powerful representation ability and fast inference speed of neural networks have enabled deep learning methods to achieve significant fluid simulations and flow predictions. According to the training strategy, these methods can be broadly divided into two categories: data-driven methods and physics-informed methods. Data-driven methods[1–6] rely on vast datasets generated by numerical solvers for their end-to-end training. The model's prediction performance is strictly limited by the results produced by the numerical solver. Additionally, the model's generalization ability is seriously affected by the distribution of labeled data, which hinders the application of such models in real-world scenarios. Since Raissi et al.[7]

proposed the unsupervised physics-informed neural network (PINN) framework, this approach has been widely employed for solving PDEs or inverse problems relating to flow fields.[8–11] PINNs use the automatic differentiation (AD) technique within the deep learning framework[12,13] to calculate spatiotemporal partial derivatives analytically. The PDE residuals are treated as loss functions, which provide the penalties through which the network is trained.

Different from the classic PINN using AD to formulate the PDE loss, some methods have employed FD-based numerical differentiation to discretize the network's output domain.[14–16] This physics-informed approach provides efficient flow predictions in parameterized flow configurations (varying geometries or boundary conditions). The FD-based methods mostly employ a convolutional neural network (CNN) as the backbone model. To capture the physical phenomena precisely, especially near wall boundaries, body-fitted multi-block-structured grids are often used in CFD. However, CNNs cannot be directly applied on irregular domains. Figure 1(a) depicts the general routine of FD methods combined with CNNs. The main drawback of this approach is that the input samples are restricted to a rectangular space (e.g., 200 × 200) in one batch, which severely limits the model's generalization ability to different geometries. Additionally, some CNN-based methods[11,14] rasterize the input mesh to a uniform Cartesian-like domain, which reduces the boundary grid resolution. Figure 2 shows the difference between the rasterization (image-like grid) and body-fitted block-structured grid. Gao et al.[16] proposed the CNN-based PhyGeoNet to solve parameterized PDEs on irregular domains. In their work, the FD method is implemented by convolutions in the uniform computational space generated by a coordinate transformation.[17–20] Although this is capable of handling irregular domains in parameterized PDE settings, it can only be trained on irregular single-block-structured grids. Thus, when solving flow cases with new geometries, PhyGeoNet has to be retrained. It is difficult for PhyGeoNet to tackle complex cases such as pipe flows with cylindrical obstacles. When FD is used to obtain numerical solutions, the physical domain is always discretized to a body-fitted block-structured grid, which creates difficulties for CNN-based methods like PhyGeoNet. Figure 3 shows a multi-block-structured grid diagram for a cylinder pipe flow, illustrating why training on this kind of body-fitted block-structured grid is challenging for PhyGeoNet. Given the geometry in Fig. 3, the convolution operation in the CNN cannot handle different grid blocks in parallel. Thus, a more general and efficient differentiable FD design is needed for such cases.

Benefiting from strong geometric adaptability, graph networks (GNs)[21] have successfully been used for flow predictions and PDE-solving tasks in a data-driven manner.[1,2,6,22–24] The inputs produced by different geometries can be concatenated and organized in an unstructured tensor form, as depicted in Figs. 1 and 3. Although block-structured grids can be easily treated as graphs and fed to the GN, a differentiable FD method for the unstructured GN output is still worth exploring. In this paper, an unsupervised FD-based physics-constrained GN framework (FDGN) is proposed, in which the GN serves as a feature extractor on flow fields discretized by block-structured grids. A differentiable graph convolution-based FD method with a grid coordinate transformation is proposed to achieve the FD-based physics constraint on irregular domains. The FD scheme in CNN-based methods is implemented with a fixed convolution kernel, which is limited to the rectangular space. In contrast, as shown in Fig. 1(b), the proposed GC-FDM can be viewed as an unlearnable spatial graph convolution and is applicable to unstructured graph outputs, which are not limited by the domain geometry. To the best of our knowledge, the proposed FDGN is the first framework to combine GN with FD and predict flows on block-structured grids.

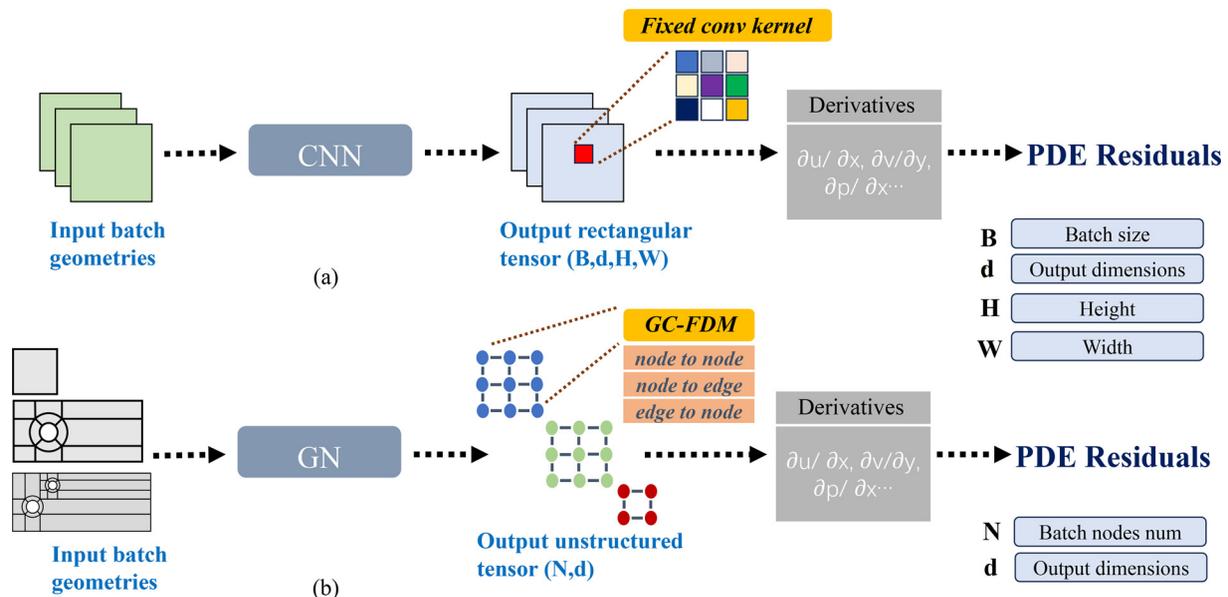

**FIG. 1.** Comparison between CNN-based FD methods (a) and the proposed finite-difference-informed graph network FDGN (b). (a) The spatial derivatives are approximated with a fixed convolutional kernel. (b) The proposed FDGN with the GC-FDM to obtain derivatives from the perspective of a graph, which are not limited by inputs with different shapes.

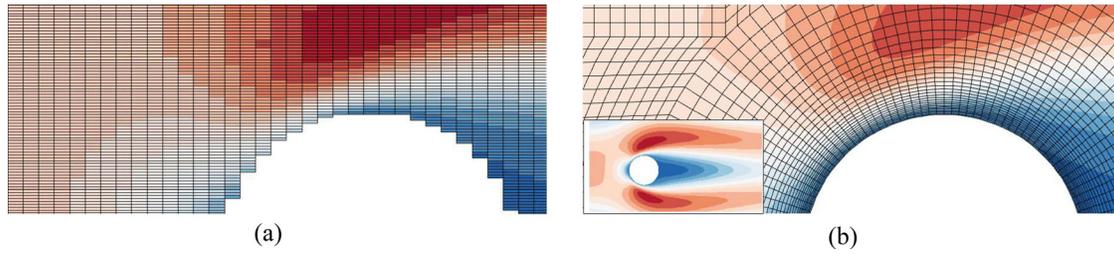

FIG. 2. (a) Flow field rasterized to the uniform Cartesian grid, which lacks the high resolution around the boundary of the body-fitted grid. The drawbacks of this method are discussed in detail in Ref. 16. (b) Magnified view of flow field solved by the proposed method on the block-structured grid.

## A. GN-based flow predictions

The strong adaptability of learning in non-Euclidean spatial domains has led to the widespread use of GN models in handling scientific problems including PDEs or flow predictions. Sanchez-Gonzalez et al.[24] proposed the GNS framework, which learns the complex interactions among fluid particles via message passing on graphs. Pfaff et al.[6] developed Meshgraphnets, which efficiently simulate a wide range of physical systems. As an improvement over GNS, Meshgraphnets encode the relative grid position as the edge feature, which significantly improves the physical simulation performance on both Lagrange and Euler systems. To overcome the problem of oversmoothing,[25] Fortunato et al.[22] introduced the multigrid scheme into Meshgraphnets. This helps the model learn and generate grid representations across different scales. Combining a GN with a differentiable CFD solver, CFD-GCN[1] achieves fast and accurate flow prediction on coarsened grids. Han et al.[23] employed the temporal attention mechanism to reduce the error accumulation in next-step prediction models, and learned the spatial flow representations using a graph autoencoder. Brandstetter et al.[2] proposed the message passing PDE solver, which encodes the PDE parameters as node features and greatly improves the generalization between different PDEs. However, the above-mentioned GN-based methods are all trained in a data-driven manner, and thus require large amounts of labeled data generated by CFD solvers.

## B. Physics-informed flow predictions

Since Raissi et al.[7] first proposed PINNs, there has been a trend to incorporate fully connected neural networks (FCNNs) and AD. To predict incompressible laminar flows, Rao et al.[8] developed a hybrid-variable PINN scheme that reduces the computational burden of calculating second-order derivatives. This scheme employs the general continuum equations together with the material constitutive law, rather than the derived Navier–Stokes equations. Jin et al.[9] proposed NSFnets for solving Navier–Stokes equations based on two different equation forms and derived a new scheme to balance the different terms in the loss function. To achieve a continuous physical representation, Spline-PINN[11] solves the incompressible Navier–Stokes equations as well as the damped wave equation. This model was trained to predict the coefficients of Hermitian splines and achieves high-accuracy continuous interpolation within a grid unit. Chiu et al.[26] established a coupled AD/FD method to improve the training efficiency of PINNs, and showed that this achieves superior performance on several challenging problems. Tang et al.[27] introduced interpolation polynomials into PINNs to solve nonlinear PDEs, and optimized the network to learn the coefficients of a power series, similar to Ref. 11. The NNfoil technique[28] solves the subsonic flow around an airfoil using a PINN, with a coordinate transformation applied so that learning is conducted in the uniform computational space instead of the physical space. Despite its remarkable performance, the use of AD in NNfoil leads to a high computational cost due to the chain rule in the computational graph, especially when calculating high-order derivatives.

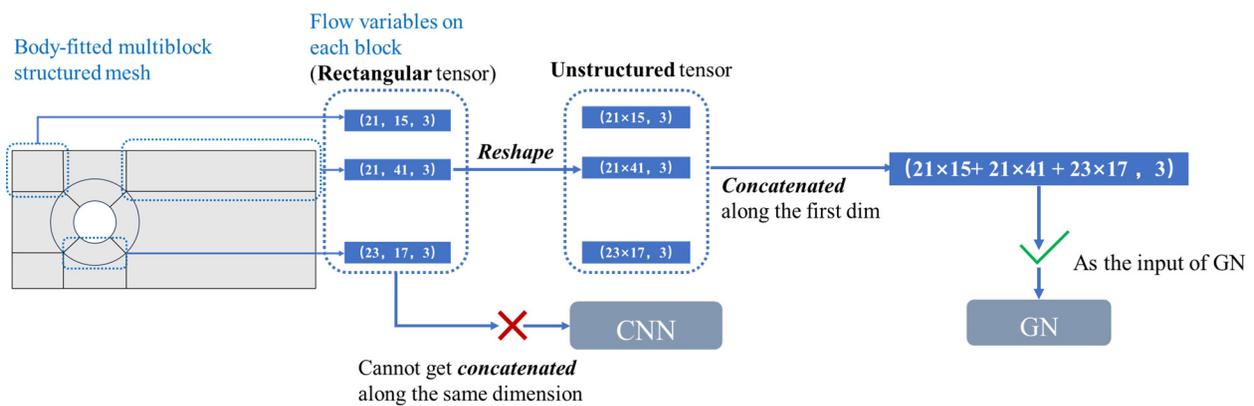

FIG. 3. Diagram of a block-structured grid (actually 12 blocks) for pipe flow. Predicting this flow is challenging for CNN-based methods because the variable tensors on different blocks or geometries are not concatenated along the same dimension as the CNN's input.

Currently, most FD-based methods use CNNs as the backbone and FD to approximate the PDE loss (residuals). The convolutional kernel with fixed parameters serves as the difference operator. Compared with FCNNs, CNNs are better at addressing grid-like data formats through the use of local parameter sharing. Wandel et al.[14] introduced the FD method with a staggered marker-and-cell grid and employed the U-Net[29] to simulate incompressible Navier–Stokes equations. This MAC-based method generates accurate and divergence-free velocity fields. The same authors also developed a MAC-based method for 3D fluid scenarios.[15] However, these two methods use a rasterized grid as the network input, and so they cannot handle a body-fitted grid. To achieve PDE solutions on irregular domains, Gao et al.[16] proposed the CNN-based PhyGeoNet, which introduced convolutions with a coordinate transformation in the computational space. This was the first CNN-based method dedicated to non-rectangular geometries and nonuniform grids. Lim et al.[30] directly replaced AD in the PINN framework with FD on 2D uniform grids. However, the above-mentioned FD-based methods are limited to single-block grids or image-like uniform grids, and cannot be applied to block-structured grids.

This study is roughly divided into four parts: Sec. II introduces the methodology of the proposed FDGN, including the problem setup, coordinate transformation, network model, and GC-FDM for physics-constrained training. Section III details the numerical results, comparing the performance of FDGN with traditional CFD methods and PINNs across various flow cases, including lid-driven cavity flow and flow around single and double cylinders. The advantages of FDGN in parameterized settings and multi-block-structured grids are also highlighted. Section IV demonstrates the limitations of the present method and presents the discussions and conclusions, summarizing the key findings and potential future work, such as extending FDGN to unsteady and even turbulent flows.

## II. PROBLEM SETUP AND METHODOLOGY

In this section, the 2D steady incompressible Navier–Stokes equations and the grid coordinate transformation are presented. Then, the GN model used for flow predictions is introduced. Furthermore, the proposed GC-FDM with a physics-constrained loss based on PDE residuals is demonstrated. Finally, the training strategy is detailed.

### A. Two-dimensional steady-state incompressible Navier–Stokes equations

Most fluid flows are governed by the Navier–Stokes equations. In this work, the 2D steady Navier–Stokes equations are considered for an incompressible flow. The dimensionless equations in the physical space $\Omega_p$ are

$$\nabla \cdot \mathbf{u} = 0 \quad \text{in} \quad \Omega_p, \tag{1}$$

$$\nabla \cdot (\mathbf{uu}) + \nabla p - \frac{1}{Re}\nabla^2 \mathbf{u} = 0 \quad \text{in} \quad \Omega_p, \tag{2}$$

where $\mathbf{u} = (u, v)$ is the velocity vector, $p$ is the pressure, and $Re$ is the Reynolds number, which is dimensionless and reflects the ratio of inertial force to viscous force in the fluid. $Re$ is defined as follows:

$$Re = \frac{U_{mean} D}{\nu}, \tag{3}$$

where $U_{mean}$ is the mean velocity, $D$ is the characteristic length of the flow, and $\nu$ is the kinematic viscosity. Equation (1) ensures that the fluid is incompressible and that the velocity field u is divergence-free. Equation (2) describes the relationship between the momentum of fluid particles and external forces.

Similar to the MAC-method,[14] the above-mentioned equations are to be solved by the GN with fixed input field variables (u and $p$) and specific Dirichlet boundary conditions $\mathbf{u}_D$,

$$\mathbf{u} = \mathbf{u}_D \quad \text{in} \quad \partial\Omega, \tag{4}$$

where $\partial\Omega$ is the flow boundary.

### B. Coordinate transformation

The FD scheme is used to obtain the PDE residuals on an irregular domain. However, this scheme cannot be directly applied on grids with curvilinear coordinates, where stretching and rotation exist. Thus, a coordinate transformation[18] from the physical space to the computational space[31–33] is required before the FD scheme can be applied.

When the equations are to be solved in curvilinear coordinates using FD, a 2D coordinate transformation from the physical space $(x, y)$ to the computational space $(\xi, \eta)$ is needed,

$$\begin{cases} \xi = \xi(x, y), \\ \eta = \eta(x, y), \end{cases} \tag{5}$$

where $\xi(\cdot)$ and $\eta(\cdot)$ denote the coordinate mapping. The inverse of the grid Jacobian, $J^{-1}$, can then be calculated as follows:

$$J^{-1} = \begin{vmatrix} x_\xi & y_\xi \\ x_\eta & y_\eta \end{vmatrix} = \begin{vmatrix} \xi_x & \eta_x \\ \xi_y & \eta_y \end{vmatrix}^{-1}, \tag{6}$$

where $\xi_x$ denotes $\frac{\partial \xi}{\partial x}$, and similarly for other variables. The derivatives $x_\xi, y_\xi, x_\eta,$ and $y_\eta$ are approximated using the second-order central difference scheme. The grid metrics $\xi_x, \xi_y, \eta_x,$ and $\eta_y$ can be obtained as follows:

$$\begin{cases} \xi_x = Jy_\eta, \\ \xi_y = -Jx_\eta, \\ \eta_x = -Jy_\xi, \\ \eta_y = Jx_\xi, \end{cases} \tag{7}$$

$$T^{ab} = \begin{bmatrix} T^{11} & T^{12} \\ T^{21} & T^{22} \end{bmatrix}$$
$$= \begin{bmatrix} (\xi_x^2 + \xi_y^2)J^{-1} & (\xi_x\eta_x + \xi_y\eta_y)J^{-1} \\ (\xi_x\eta_x + \xi_y\eta_y)J^{-1} & (\eta_x^2 + \eta_y^2)J^{-1} \end{bmatrix}, \tag{8}$$

and the covariant velocities as

$$U = (u\xi_x + v\xi_y)J^{-1}, \quad V = (u\eta_x + v\eta_y)J^{-1}. \tag{9}$$

After applying the mapping in Eq. (5), the 2D steady Navier–Stokes equations in the physical space are transformed into the general form in the computational space,

$$\underbrace{\frac{\partial E}{\partial \xi} + \frac{\partial F}{\partial \eta}}_{\text{Inviscid flux term}} - \underbrace{\frac{\partial E_v}{\partial \xi} - \frac{\partial F_v}{\partial \eta}}_{\text{Viscous flux term}} = 0, \quad (10)$$

where the inviscid fluxes $E$ and $F$ in the transformed space are written as follows:

$$E = \begin{bmatrix} U \\ uU + p\xi_x J^{-1} \\ vU + p\xi_y J^{-1} \end{bmatrix}, \quad F = \begin{bmatrix} V \\ uV + p\eta_x J^{-1} \\ vV + p\eta_y J^{-1} \end{bmatrix}, \quad (11)$$

and the viscous fluxes $E_v$ and $F_v$ are

$$E_v = \frac{1}{Re}\begin{bmatrix} 0 \\ T^{11}\frac{\partial u}{\partial \xi} + T^{12}\frac{\partial u}{\partial \eta} \\ T^{11}\frac{\partial v}{\partial \xi} + T^{12}\frac{\partial v}{\partial \eta} \end{bmatrix}, \quad F_v = \frac{1}{Re}\begin{bmatrix} 0 \\ T^{21}\frac{\partial u}{\partial \xi} + T^{22}\frac{\partial u}{\partial \eta} \\ T^{21}\frac{\partial v}{\partial \xi} + T^{22}\frac{\partial v}{\partial \eta} \end{bmatrix}. \quad (12)$$

The grid metrics $\xi_x, \xi_y, \eta_x, \eta_y$ and the Jacobian $J$ are treated as constants and moved into the partial derivative in a flux form. Thus, the FD scheme can be simply applied to the fluxes $E, F, E_v,$ and $F_v$ along the $\xi$ and $\eta$ directions. Similar to PhyGeoNet,[16] the second-order derivative is not completely expanded according to the chain rule, which helps avoid the overcomplex FD approximation.

### C. Network model

The graph model proposed in Refs. 34 and 35 is utilized, which has an encoder–processor–decoder structure similar to that of Meshgraphnet,[6] but with a few modifications. The input grid is represented by a directed graph $G = (V, E)$ with node set $V$ and edge set $E$. The graph contains node features $\mathbf{X} = \{x_r \in \mathbb{R}^{c1} : r \in V\}$ and edge features $\mathbf{E} = \{e_{r,s} \in \mathbb{R}^{c2} : (r,s) \in E\}$. The node features of node $r$ are initialized with input flow variables $\mathbf{f}_r = [u_r, v_r, p_r]$ and the one-hot vector $\mathbf{n}$, which indicates the node type. The relative grid position $z_r - z_s$, its norm $\|z_r - z_s\|$, and the flow field differences $\mathbf{f}_r - \mathbf{f}_s$ are encoded as the input edge features for edge $e_{r,s}$.

#### 1. Encoder

The encoder projects the input node and edge features to the $d$-dimensional latent space using the mappings $\phi_{E-n}$ and $\phi_{E-e}$:

$$\begin{aligned} x_r &= \phi_{E-n}([\mathbf{f}_r, \mathbf{n}]), \\ e_{r,s} &= \phi_{E-e}([z_r - z_s, \|z_r - z_s\|, \mathbf{f}_r - \mathbf{f}_s]). \end{aligned} \quad (13)$$

Both $\phi_{E-n}$ and $\phi_{E-e}$ are implemented using two-layer multilayer perceptrons (MLPs) with SiLU activation.[36] Layernorm is employed to obtain the final output.

#### 2. Processor

The processor unit takes the projected $x_r$ and $e_{r,s}$ as its input, which is composed of $K$ message passing blocks. Each block is applied in sequence to the output of the previous block. Different from Meshgraphnets, the $i$th grid cell features $c'_i$ are calculated and used to update the node features as follows:

$$\begin{aligned} e'_{r,s} &= \phi_{P-e}(e_{r,s}, x_r, x_s), \\ c'_i &= \frac{1}{N_e}\left(\sum_{e \in cell_i} e'_{r,s}\right), \\ x'_r &= \phi_{P-n}\left(x_r, \frac{1}{N_c}\sum c'_i\right), \end{aligned} \quad (14)$$

where $N_e$ denotes the number of edges of a cell and $N_c$ is the number of cells sharing node $r$. The MLPs used to learn the mappings $\phi_{P-e}$ and $\phi_{P-n}$ have the same structure as those in the encoder. The node features are updated according to the neighboring cells rather than edges, unlike in the original Meshgraphnet. The node features beyond a one-hop distance are aggregated into cells, which encourages more efficient message passing. More details about this modified grid cell-based message passing can be found in Ref. 34.

#### 3. Decoder

The decoder module decodes the node representation output by the $K$th message passing block in the processor to the flow variables $u$, $v$, and $p$. The decoder mapping $\phi_D$ is also implemented using a two-layer MLP with SiLU activation, but without Layernorm. The output $u$, $v$, and $p$ are used in the GC-FDM (see Sec. II D) to obtain the PDE residuals (loss) for network training.

### D. GC-FDM

#### 1. Domain transformation

In traditional CFD, a body-fitted block-structured grid with a coordinate transformation[19] is used to simulate complex geometries. The physical domain is discretized into multiple blocks that are connected by an interface where adjacent blocks share the same grid nodes. Each block is represented by an independent curvilinear coordinate system. The FD scheme with a coordinate transformation is employed to approximate the solution block by block in the computational domain.

Inspired by this, the grid block is treated as the basic unit when conducting the GC-FDM. After obtaining the decoded variables (predicted $u$, $v$, $p$) on the output graph in the physical space $G_{phy} = (V_{phy}, E_{phy})$, the original $G_{phy}$ is converted into a new graph $G_{com} = (V_{com}, E_{com})$ that belongs to the computational space. Figure 4 illustrates the above-mentioned transformation process. For simplification, magnified grids are used rather than the whole domain to represent $G_{phy}$ and $G_{com}$. In $G_{com}$, the blocks are separated at each interface and the node features are repeated along the interface. This change in topology requires the node indices in $G_{com}$ to be rearranged according to the number of nodes (features). Let $X_{phy} \in \mathbb{R}^{|V_{phy}|\times 3}$ denote the feature matrix of $G_{phy}$ and $X_{com} \in \mathbb{R}^{|V_{com}|\times 3}$ denote the feature matrix of $G_{com}$. By selecting the tensor index along the first dimension, $X_{com}$ can be obtained by

$$X_{com} = X_{phy}[index\_block], \quad (15)$$

where $index\_block \in \mathbb{R}^{|V_{com}|}$ stores the node indices of all blocks from $G_{phy}$, which are not yet rearranged. The node features are shared between the blocks on each interface. Suppose that the whole geometry contains $I$ blocks and $|V_i|$ represents the number of nodes in the $i$th block. It can be inferred that

$$|V_{com}| = \left(\sum_{i=1}^{I} |V_i|\right) \geq |V_{phy}|. \quad (16)$$

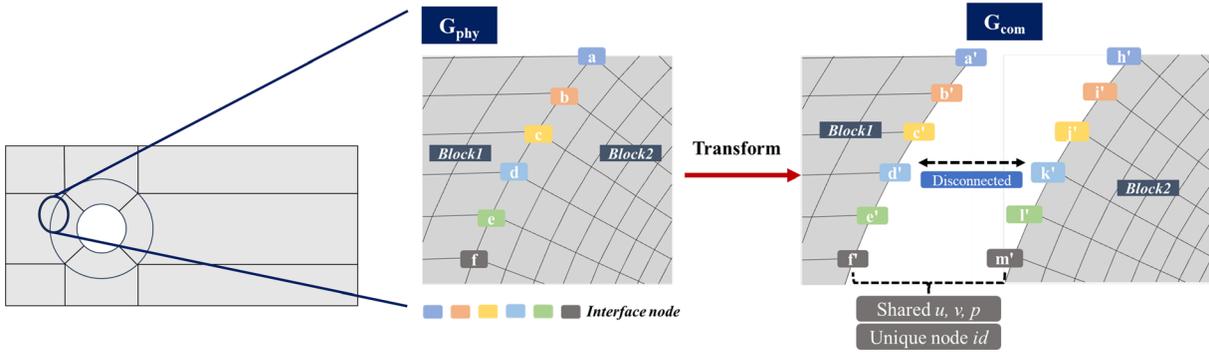

**FIG. 4.** The left grid is modeled as $G_{phy}$, the right one is $G_{com}$, and the letters denote different node indices. The overall topology differs between these two kinds of graph, with $G_{com}$ containing more nodes and edges because of node feature sharing between block pairs. The nodes in the same colors share the same predicted features ($u$, $v$, $p$). The node indices in $G_{com}$ are rearranged so that each node index is unique.

Then, the $E_i \in \mathbb{R}^{2 \times |E_i|}$ is denoted as the edge set of each block, which is composed of the rearranged node indices in $G_{com}$. The edge set $E_{com}$ can be obtained using the *concatenate* operation as follows:

$$E_{com} = concatenate(E_1, E_2, \ldots, E_I). \quad (17)$$

Similarly, the $|E_{com}|$ satisfies

$$|E_{com}| = \left(\sum_{i=1}^{I} |E_i|\right) \geq |E_{phy}|. \quad (18)$$

In the one-block scenario ($I = 1$), $|V_{com}| = |V_{phy}|$ and $|E_{com}| = |E_{phy}|$ because there is no block separation and node repetition. This domain transformation ensures communication among the flow variables in adjacent blocks, as in traditional CFD.

With the selected node features $X_{com}$ and adjacencies $E_{com}$, the $G_{com}$ is constructed via this transformation. Although the adjacent blocks are no longer connected, the topology within each block is retained, which is vital in mimicking the traditional block-based FD scheme in CFD. Then, the GC-FDM is applied to the new $G_{com}$, which is equivalent to conducting FD among blocks simultaneously in a differentiable way.

### 2. FD scheme

Before describing the implementation of GC-FDM, the finite-difference scheme is introduced, based on which GC-FDM is designed. The discretization uses a three-point stencil (Fig. 5) along the $\xi$ or $\eta$ direction. Figure 5 concisely depicts this in curvilinear coordinates. Based on this stencil, the inviscid and viscous flux terms in Eqs. (11) and (12) can be approximated using the FD scheme.

To discretize the inviscid flux $E$ and $F$, the second-order central difference scheme are used to obtain the corresponding derivatives as follows:

$$\begin{cases} \left(\dfrac{\partial E}{\partial \xi}\right)_{i,j} = (E_{i+1,j} - E_{i-1,j})/2, \\ \left(\dfrac{\partial F}{\partial \eta}\right)_{i,j} = (F_{i,j+1} - F_{i,j-1})/2. \end{cases} \quad (19)$$

To discretize the viscous flux $E_v$ and $F_v$, the half points are introduced to each edge. The overall process can be divided into two steps. In the first step, $E_v$ and $F_v$ to the half point are projected to the half point by averaging adjacent nodes,

$$\begin{cases} (E_v)_{i\pm 1/2,j} = ((E_v)_{i\pm 1,j} + (E_v)_{i,j})/2, \\ (F_v)_{i,j\pm 1/2} = ((F_v)_{i,j\pm 1} + (F_v)_{i,j})/2. \end{cases} \quad (20)$$

In the second step, the FD scheme is applied to the half points to obtain the derivatives of $E_v$ and $F_v$ on node $(i,j)$,

$$\begin{cases} \left(\dfrac{\partial E_v}{\partial \xi}\right)_{i,j} = (E_v)_{i+1/2,j} - (E_v)_{i-1/2,j}, \\ \left(\dfrac{\partial F_v}{\partial \eta}\right)_{i,j} = (F_v)_{i,j+1/2} - (F_v)_{i,j-1/2}. \end{cases} \quad (21)$$

Using the above-mentioned derivatives, the residuals of Eq. (10) are calculated on node $(i,j)$.

### 3. Bridging the gap between FD and graph convolution

The goal is to learn the flow representations on block-structured grids, which are represented by graphs. However, there is still a gap between the FD implementation and the unstructured graph data.

Since Gilmer et al.[37] first proposed the message passing neural network, most spatial graph convolution methods[38–41] have been interpreted or generalized using this framework. According to Ref. 42, the message passing mechanism can be divided into two steps: a message aggregate step and a feature update step. In the message aggregate step, the information from $\mathcal{N}(r)$ (neighbors of node $r$) is aggregated into a message vector $m_r$, which can be formulated as follows:

$$m_r = A(\phi(x_r^{old}, x_s^{old}, e_{s,r}), s \in \mathcal{N}(r)), \quad (22)$$

where $x_r^{old} \in \mathbb{R}^F$ denotes the node feature of node $r$ before updating, $e_{s,r} \in \mathbb{R}^D$ denotes the edge features from node $s$ to node $r$, and $\phi$ is a customized mapping function. The operator $A(\cdot)$ is a differentiable, permutation-invariant function (e.g., sum, mean, or maximum). After obtaining the messages from neighbor nodes, the feature update step

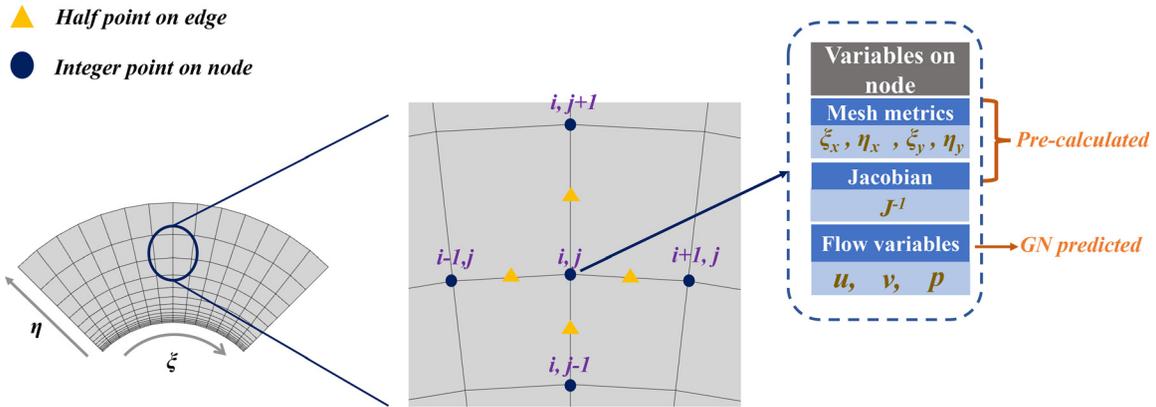

**FIG. 5.** Diagram of the three-point stencil. To approximate the derivative at node $(i, j)$, information from three points (the node itself and two adjacent nodes along the $\xi$ or $\eta$ direction) is used, with the edge between the two adjacent nodes storing variables at half points. The grid metrics and Jacobian are unlearnable per-node variables, which are pre-calculated in this curvilinear coordinate system.

updates the representation of node $r$ with the message, which can be described as follows:

$$x_r^{new} = \gamma(x_r^{old}, m_r), \quad (23)$$

where $\gamma^l(\cdot)$ denotes the update mapping, which can also be customized.

As the message passing framework models the relations between different entities[21] (i.e., node-to-node, node-to-edge, and edge-to-node), The per-node derivative approximation is demonstrated in Eqs. (19)–(21) can be modeled by message passing in the three-point stencil. Node $(i, j)$ serves as target node $r$, where the corresponding derivatives are demanded. As shown in Fig. 6, the neighbors $\mathcal{N}(r)$ are

$$\{node(i-1, j), node(i+1, j), node(i, j-1), node(i, j+1)\}. \quad (24)$$

The relationship between the difference scheme and graph convolution is now illustrated.

*a. Node-to-node message passing for approximating* $\left(\frac{\partial E}{\partial \xi}\right)_{ij}$ *and* $\left(\frac{\partial F}{\partial \eta}\right)_{ij}$. Figure 6(a) shows that the central difference scheme in Eq. (19) can be achieved using node-to-node message passing. First, the message vector $m_r$ of node $(i, j)$ is obtained through the following matrix multiplication:

$$m_{(i,j)} = \begin{bmatrix} -\frac{1}{2}, \frac{1}{2} \end{bmatrix}^T \begin{bmatrix} E_{i-1,j}, E_{i+1,j} \\ F_{i,j-1}, F_{i,j+1} \end{bmatrix}, \quad (25)$$

where the target node feature before updating $x_{(i,j)}^{old} = 0$ and edge feature $e_{s,r} = 0$; the inviscid fluxes $E$ and $F$ represent neighboring node features. As each neighboring node is multiplied by a fixed weight and aggregated via summation, the unlearnable linear transformation $[-\frac{1}{2}, \frac{1}{2}]^T$ is equivalent to mapping $\phi(\cdot)$ with the permutation invariant function $A(\cdot)$ in Eq. (22).

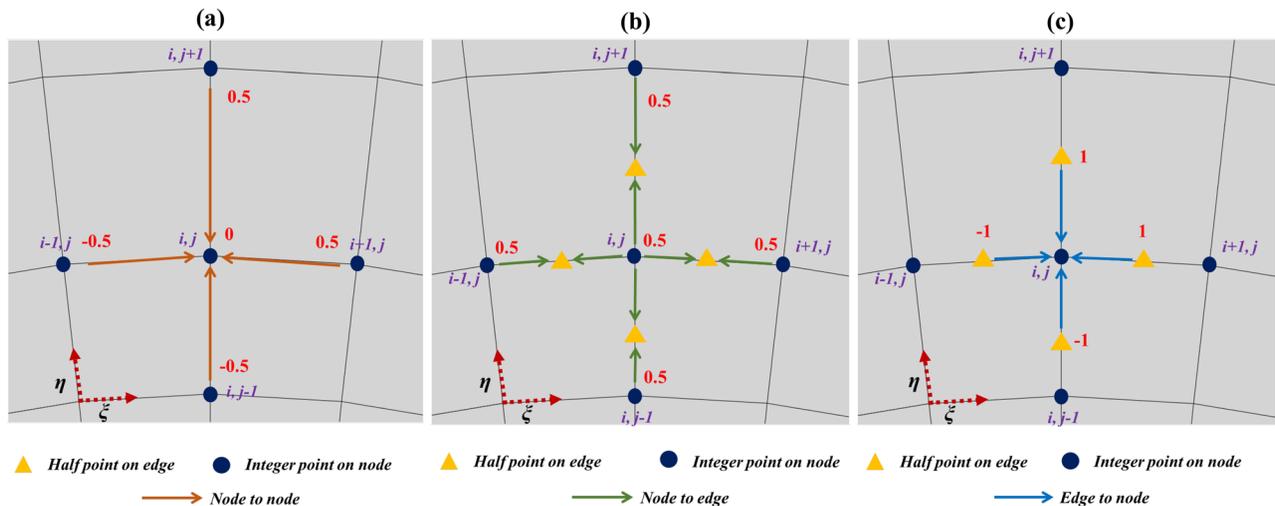

**FIG. 6.** Diagram of the FD scheme represented by graph convolutions. The derivative on node $(i, j)$ is approximated using information from three points (the node itself and two adjacent nodes along the $\xi$ or $\eta$ direction), and the edge between the two adjacent nodes stores the variables at half points. The grid metrics and Jacobian are unlearnable per-node variables, which are pre-calculated in this curvilinear coordinate system. (a) Conducting central difference via node-to-node. (b) Projecting variables to half points via node-to-edge. (c) Approximating the viscous term via edge-to-node.

In the feature update step, $\gamma(\cdot)$ can be defined as follows:

$$x_{(i,j)}^{new} = \gamma(x_{(i,j)}^{old}, m_{(i,j)}) = 0 \cdot x_{(i,j)}^{old} + m_{(i,j)}. \quad (26)$$

Thus, the updated feature $x_{(i,j)}^{new}$ in Eq. (23) is exactly the derivative on node $(i,j)$,

$$x_{(i,j)}^{new} = \left[ \left(\frac{\partial E}{\partial \xi}\right)_{i,j}, \left(\frac{\partial F}{\partial \eta}\right)_{i,j} \right]$$
$$= \left[\frac{(E_{i+1,j} - E_{i-1,j})}{2}, \frac{(F_{i,j+1} - F_{i,j-1})}{2}\right]. \quad (27)$$

b. *Node-to-edge-to-node message passing for approximating* $\left(\frac{\partial E_v}{\partial \xi}\right)_{i,j}$ *and* $\left(\frac{\partial F_v}{\partial \eta}\right)_{i,j}$. The two steps of Eqs. (20)–(21) in the approximation of $\left(\frac{\partial E_v}{\partial \xi}\right)_{i,j}$ and $\left(\frac{\partial F_v}{\partial \eta}\right)_{i,j}$ can also be generalized by message passing. The projection to half points is achieved in a node-to-edge manner, as shown in Fig. 6(b), and the edge feature $e_{s,r}$ in Eq. (22) is calculated by

$$\begin{bmatrix} e_{(i-1,j),(i,j)} \\ e_{(i+1,j),(i,j)} \\ e_{(i,j-1),(i,j)} \\ e_{(i,j+1),(i,j)} \end{bmatrix} = \left[\frac{1}{2}, \frac{1}{2}\right]^T \begin{bmatrix} (E_v)_{i-1,j}, (E_v)_{i,j} \\ (E_v)_{i+1,j}, (E_v)_{i,j} \\ (F_v)_{i,j-1}, (F_v)_{i,j} \\ (F_v)_{i,j+1}, (F_v)_{i,j} \end{bmatrix}. \quad (28)$$

Next, message $m_{(i,j)}$ is calculated using the linear transformations $[-1, 1, 0, 0]^T$ and $[0, 0, -1, 1]^T$

$$m_{(i,j)} = \left[ \begin{bmatrix} -1, & 1, & 0, & 0 \end{bmatrix}^T \begin{bmatrix} e_{(i-1,j),(i,j)} \\ e_{(i+1,j),(i,j)} \\ e_{(i,j-1),(i,j)} \\ e_{(i,j+1),(i,j)} \end{bmatrix} \right.$$
$$\left. \times \begin{bmatrix} 0, & 0, & -1, & 1 \end{bmatrix}^T \begin{bmatrix} e_{(i-1,j),(i,j)} \\ e_{(i+1,j),(i,j)} \\ e_{(i,j-1),(i,j)} \\ e_{(i,j+1),(i,j)} \end{bmatrix} \right], \quad (29)$$

the above-mentioned process can be viewed as an edge-to-node message aggregation, as illustrated in Fig. 6(c). Similarly, using the update function $\gamma(\cdot)$ in Eq. (26), $\left(\frac{\partial E_v}{\partial \xi}\right)_{i,j}$ and $\left(\frac{\partial F_v}{\partial \eta}\right)_{i,j}$ are treated as the updated $x_{(i,j)}^{new}$:

$$x_{(i,j)}^{new} = \left[ \left(\frac{\partial E_v}{\partial \xi}\right)_{i,j}, \left(\frac{\partial F_v}{\partial \eta}\right)_{i,j} \right]$$
$$= [e_{(i+1,j),(i,j)} - e_{(i-1,j),(i,j)}, e_{(i,j+1),(i,j)} - e_{(i,j-1),(i,j)}]. \quad (30)$$

Analogous to CNNs, which employ fixed convolution kernels to obtain the spatial derivatives, the proposed GC-FDM utilizes the above-mentioned unlearnable linear transformations, which convert the stencil-based FD into a weighted summation in the local area. Additionally, the above-mentioned operations in GC-FDM are completely differentiable and easy to implement because the node adjacency within each block is structured.

### E. Network training

This subsection demonstrates how the GN model is trained under the FDGN framework. Figure 7 shows the overall training process. The training loop starts from the training pool, which contains different block-structured grids with corresponding flow fields. A fixed number of grids are randomly sampled to build one graph batch. As described in Sec. II C, the node and edge features are encoded to the graph. The built batch graph is fed to the GN model to obtain the predicted flow fields. After imposing boundary conditions on the output flow fields, the original flow fields in the pool will be replaced by the output ones. Meanwhile, the updated flow fields with the pre-calculated mesh metrics and Jacobians are sent to the proposed GC-FDM to approximate the derivatives in the computational space. Finally, boundary residual processing is introduced to the output residuals from GC-FDM and calculates the PDE loss to train the GN.

More details will be given for parts of the training framework that have not been described in previous sections.

#### 1. Training pool

The training pool strategy proposed by Wandel *et al.*[14] is employed with flow variables set to zero in the first iteration. The model is trained autoregressively as described in Refs. 6, 14, 34, and 35, with the variables at the current iteration $t$ taken as the input for generating the variables at the next iteration $t + 1$. In each training iteration, a fixed number of structured grids will be extracted from the training pool. With corresponding flow variables ($u^t$, $v^t$, and $p^t$) at iteration $t$ and geometry information (coordinate $z$), a graph batch is constructed as the model's input, which is depicted in Fig. 7. The predicted outputs are set as the flow fields at iteration $t + 1$ and then passed back to the training pool to replace the flow fields from the previous iteration. Although the goal is to solve the steady-state equations without time-stepping, this introduced iterative training strategy is not changed.

#### 2. Boundary condition imposing

Gao *et al.*[16] proposed the imposition of hard boundary conditions (BCs), such as Dirichlet or Neumann BCs, directly to the output flow field by tensor padding. Different from PINNs,[7] which treat the BCs as an additional penalty, the imposition of hard BCs ensures that the BCs are strictly satisfied on the domain boundary and boosts the training convergence. Similar to Gao *et al.*,[16] for Dirichlet BCs, hard BCs are imposed, albeit with a slight difference. As the GC-FDM is conducted on graphs instead of rectangular tensors[16] where padding cannot be applied, the predicted values on boundary nodes are directly replaced with the target Dirichlet BCs. In fact, a per-node mask *BC* is used that indicates the domain boundary as follows:

$$\mathbf{u}[BC] = \mathbf{u}_D, \quad (31)$$

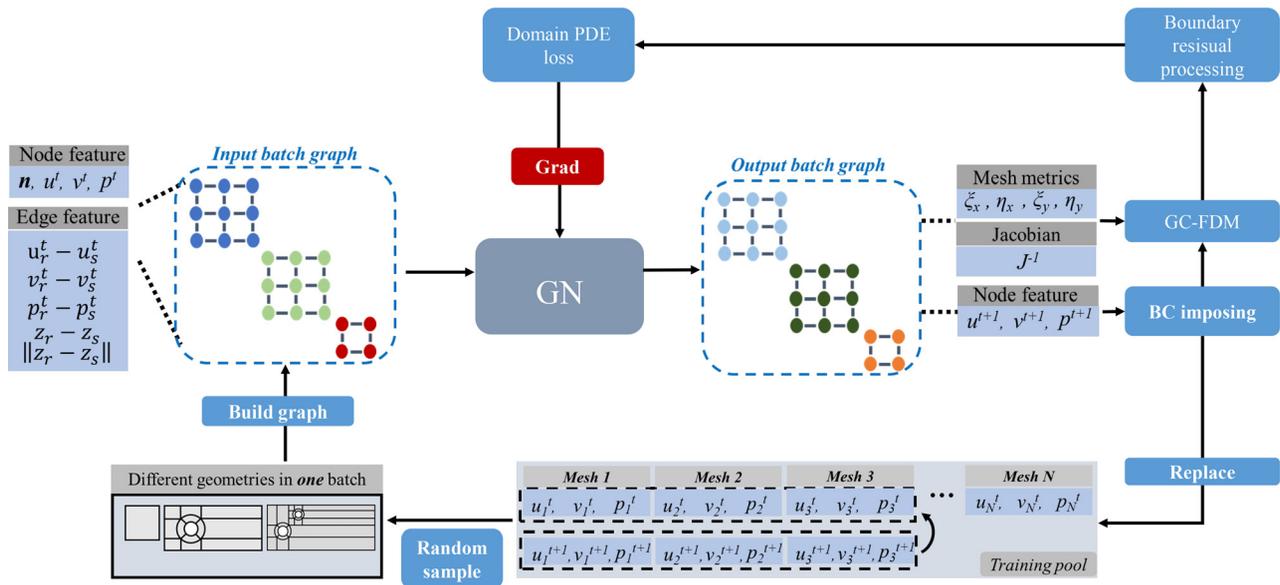

FIG. 7. Pipeline of the training framework.

where $BC = [m_1, ..., m_i, ..., m_{|V_{phy}|}]$ and $m_i \in \{True, False\}$ (True indicates a node belonging to the boundary).

The Neumann BCs for the Navier–Stokes equations are set to constrain the outlet pressure. Following Ref. 43, the Neumann BCs is defined as follows:

$$\nu \partial_n u - pn = 0, \qquad (32)$$

where $n$ represents the outlet normal vector. To impose the Neumann BCs in Eq. (32), it is treated as an additional loss penalty, as in PINNs, which is discretized by the proposed GC-FDM.

### 3. Boundary residual processing

In Sec. II D, internal nodes in a block are mainly focused on to illustrate the GC-FDM process. Here, more details about the calculation of residuals on the block boundary are given, which is introduced in Fig. 7. As shown in Fig. 8, adjacent nodes from opposite blocks are incorporated to maintain the three-point stencil when employing GC-FDM.

As Dirichlet BCs have already been imposed on the boundary nodes, the exact flow field values are obtained naturally. Thus, the residuals are not required on boundary nodes, and zero values are just assigned to these nodes, then a mask will be used to eliminate these assigned *fake* residuals. The hard imposed Dirichlet BCs are only involved in the residual approximation of internal nodes and interface nodes, similar to traditional CFD.

The residuals obtained by GC-FDM are on $G_{com}$, which means that the output residual matrix $\mathbf{R} \in \mathbb{R}^{|V_{com}|}$ due to the feature sharing on each interface. Therefore, $\mathbf{R}$ needs to be projected back to the actual physical space. Theoretically, the residuals of the interface nodes in one block and their counterparts in the opposite block should be identical because they share the same coordinates in the physical space. Hence, these residuals are averaged to obtain the per-node residual in physical space. Figure 9 demonstrates the above-mentioned residual processing. However, for the single-block geometry (a cavity), the averaging is not required because there is no interface in the grid.

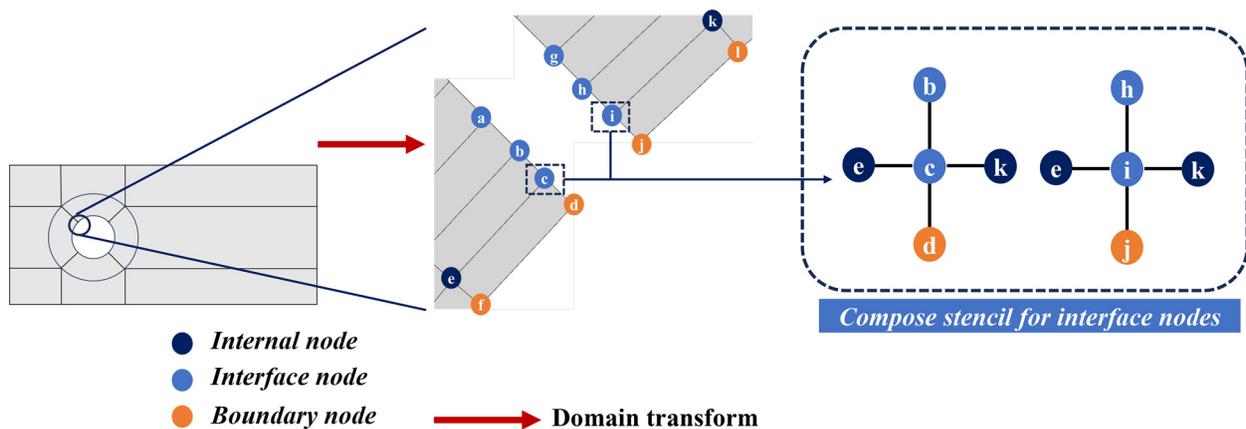

FIG. 8. Example of maintaining the three-point stencil for interface nodes. If a node is on the interface and on the flow boundary, it is classified as a flow boundary node.

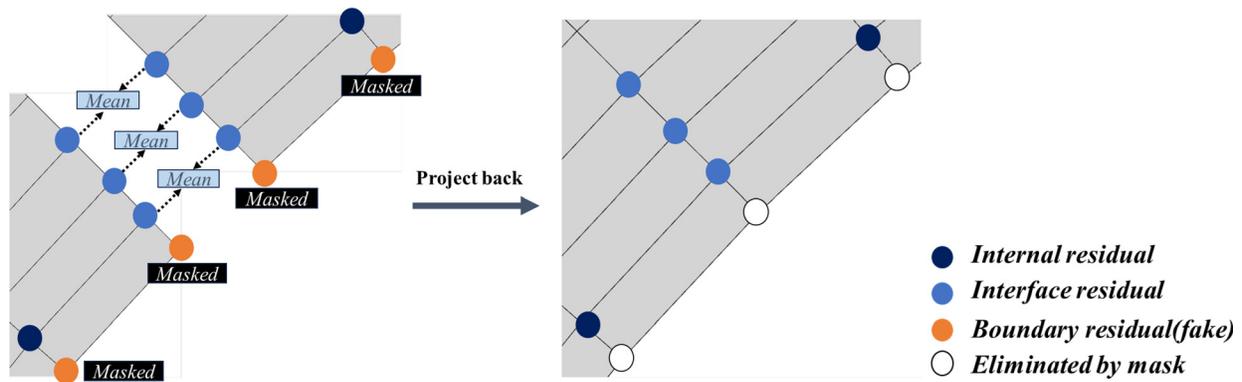

**FIG. 9.** Diagram explaining how the per-node residual of the original physical domain is obtained, which is used to train the GN.

### 4. Domain loss for network training

Finally, the domain loss function for network training is obtained. The gradients are calculated using the residuals of Eqs. (10) and (32) as shown in Fig. 7. In fact, the residuals of Eq. (10) are in a 3D vector form and can be denoted as $[R_{cont}, R_{mom-x}, R_{mom-y}]$, representing the residuals of continuity, $u$ momentum, and $v$ momentum, respectively. Then, $R_p$ is denoted as the outlet pressure residual of Eq. (32). The loss function $L$ is formulated in physical space $\Omega_p$ and on the outlet boundary $\partial\Omega_{outlet}$ as follows:

$$L = \alpha \cdot \|R_{cont}\|^2 + \beta \cdot \|R_{mom-x}\|^2 + \lambda \cdot \|R_{mom-y}\|^2 + \zeta \cdot \|R_p\|^2, \quad (33)$$

where $\alpha$, $\beta$, $\lambda$, and $\zeta$ are hyperparameters that balance the contributions of each loss term. Empirically, $\alpha$ is set to 10, $\beta$ set to 1, $\lambda$ set to 1, and $\zeta$ set to $1 \times 10^{-3}$.

## III. NUMERICAL RESULTS

In this section, the advantages of the proposed FDGN in obtaining parameterized solutions with varying geometries and boundary conditions are first studied. To evaluate the prediction performance, a finite element-based CFD solver is set as the reference for comparison. Comparisons are conducted using both trained and untrained boundary conditions to test the model's inference ability. To demonstrate the solving efficiency of the proposed method, comparisons against a classic FCNN-based PINN are conducted in a nonparameterized manner. Finally, nonparameterized solutions are obtained for the 30P30N airfoil to fully investigate the multi-block processing performance with this more challenging and complex geometry.

### A. Comparisons with CFD for parameterized solving

Typical flow cases of the Navier–Stokes equations are considered: a lid-driven cavity flow and the pipe flow around single and double cylinders. All domains are discretized by block-structured grids that vary from single blocks to multiple blocks. Details of the datasets used for the parameterized solutions are presented in Table I. To solve these various flow types with different boundary conditions, FDGN is trained once in a parameterized setting with this dataset. Different from PINNs and CNN-based methods, solutions with varying geometries and boundary conditions can be obtained simultaneously, which highlights the significant advantage of the proposed FDGN over the other two approaches.

The number of message passing blocks $K$ is set to 12 and the dimension of the latent space to 64 (these values are retained for all experiments described in this paper). During the training stage, four grid samples are randomly selected for minibatch training and 24 batches are extracted per training epoch. The maximum number of iterations $t$ in the training pool is set to 300, which means that given specific BCs, the flow field will evolve 300 times before being reset to the zero-initialized flow field with new BCs. The initial learning rate is set to $1 \times 10^{-4}$ and will be multiplied by a decay factor of $10^{-1}$ after 10 000 epochs. The AdamW[44] optimizer is employed to solve the model parameters. The model is trained for a total of 25 000 epochs to ensure good convergence. All training and testing procedures are implemented in the PyTorch framework[45] using an Nvidia RTX 4070 GPU.

**TABLE I.** Dataset statistics of the parameterized experiments on different block-structured grids and boundary conditions. The training $Re$ number is controlled within a range and a fixed sampling interval $\delta Re$. All geometries are uniformly distributed. There are 96 grids in the dataset and each geometry accounts for 1/3 of the total number of grids.

| Dataset with hybrid geometries and various boundary conditions | | | | | |
|---|---|---|---|---|---|
| Flow case | Geometry | Number of blocks and interfaces | Number of nodes | $Re$ range | $\delta Re$ |
| Lid-driven cavity flow | Cavity | 1, 0 | $55 \times 55$ | [100,400] | 100 |
| Pipe flow around a cylinder | Single cylinder | 12, 16 | 7498 | [12,36] | 8 |
| | Double cylinder | 23, 44 | 12851 | [18,30] | 6 |

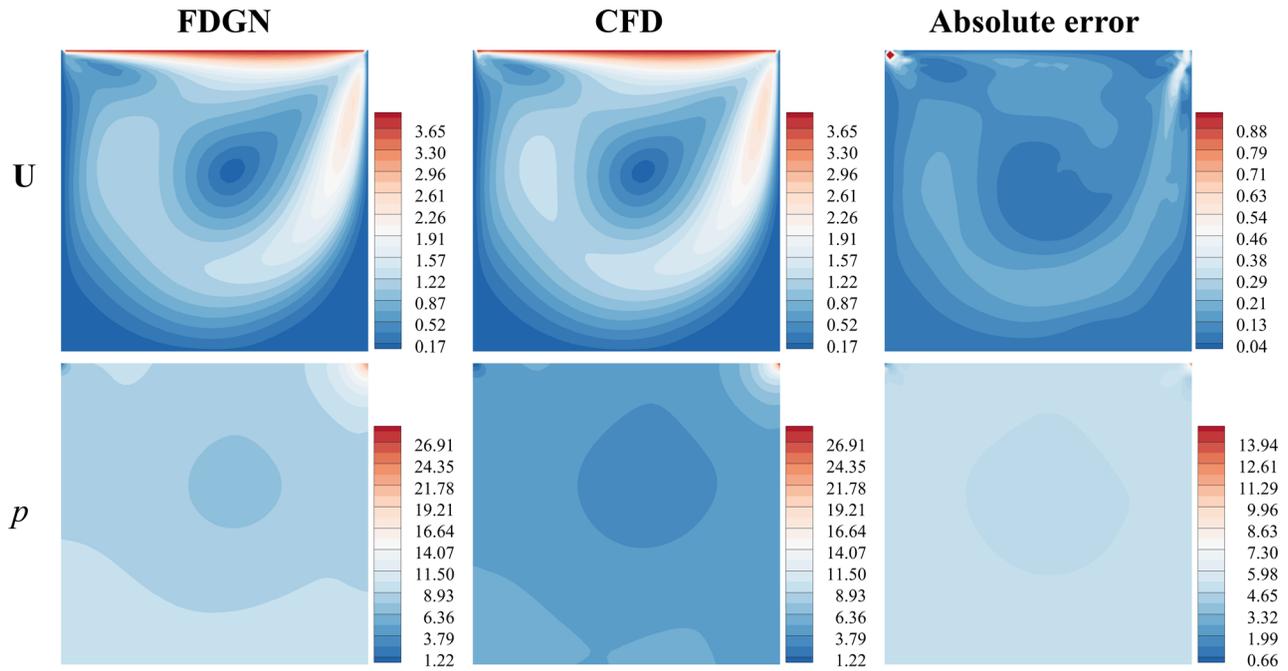

**FIG. 10.** Contour comparison at $Re = 400$ for the lid-driven cavity flow, which is seen in the hybrid dataset. The top and bottom rows show the absolute errors of the velocity magnitude $U = \sqrt{u^2 + v^2}$ and the pressure field $p$.

#### 1. Lid-driven cavity flow

The physical domain is $\Omega_{phy} = [0, 1] \times [0, 1]$. The velocity field on the wall boundaries follows the no-slip condition, except on the top wall, where the moving speed can be nonzero. For the BCs of pressure $p$, a pressure constraint is assigned in the upper left corner instead of a zero-gradient on no-slip walls. As the physical domain of this case is rectangular and there are no obstacles in the domain, one single block is suitable for the discretization. Flow structures are of great importance in evaluating the performance of the proposed method. The prediction accuracy at $Re = 400$ is visualized in Fig. 10. The proposed FDGN achieves a relatively low prediction error compared with the CFD reference.

The errors are mainly distributed in the central primary vortex region and the upper left and -right corners of the cavity. First, the flow in the primary vortex region is strongly nonlinear, and the vortex

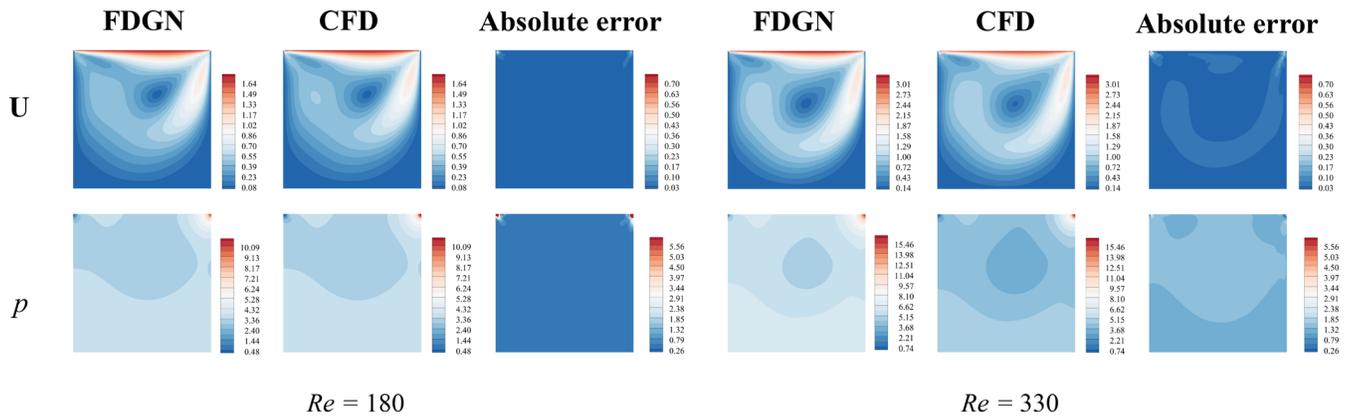

$Re = 180$          $Re = 330$

**FIG. 11.** Contour comparisons with the CFD at the unseen $Re = 180$ and $Re = 330$.

TABLE II. Relative mean absolute error for lid-driven cavity flow at various Re. The symbol ∗ denotes unseen cases outside of the hybrid dataset.

| Variable | Velocity magnitude | Pressure |
| --- | --- | --- |
| Relative error (Re = 100) | 0.011 | 0.35 |
| Relative error (Re = 180)∗ | 0.015 | 0.37 |
| Relative error (Re = 200) | 0.016 | 0.39 |
| Relative error (Re = 300) | 0.028 | 0.33 |
| Relative error (Re = 330)∗ | 0.033 | 0.30 |
| Relative error (Re = 400) | 0.037 | 0.56 |

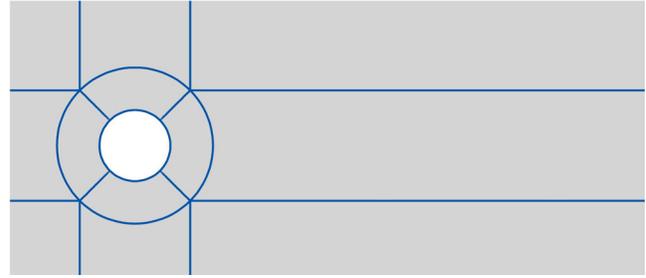

FIG. 13. Block partitioning of the structured grid for the single-cylinder case.

formation, evolution, and interaction are highly nonlinear, which increases the difficulty of predicting the flow field. In the upper-left and -right corners, the field gradients are relatively high and change rapidly. Additionally, because the corner points are located at the intersection of the top lid and walls, they are affected by different BCs at the same time, which enhances the difficulty of obtaining solutions at these corner points. The unseen BCs at $Re = 180$ and $Re = 330$ are visualized to demonstrate the model's generalization ability for the lid-driven cavity flow case. As shown in Fig. 11, the errors in the main vortex region and cavity corners increase as $Re$ becomes higher. However, the inference results still agree well with the CFD reference, which demonstrates that the model is able to learn the flow patterns through the parameterized physics-constrained training and generalize them to unseen cases.

A quantitative comparison is presented for various $Re$ including trained and untrained BCs. Table II reports the relative mean absolute error $\mathbf{e}_f = \frac{\sum |f^{FDGN} - f^{CFD}|}{\sum |f^{CFD}|}$ between FDGN and the CFD reference, where $f = U$ (velocity magnitude) or $p$ (pressure). First, the overall relative errors in the velocity fields have a low order of magnitude ($10^{-2}$), which illustrates the strong performance of FDGN. However, the errors increase with increasing $Re$ because the flow becomes more complex. In addition, the relative errors in the pressure fields are much higher than those of the velocity fields, which indicates the difficulty of predicting the pressure field in incompressible flows where the velocity and pressure fields are coupled and there is no explicit equation for the pressure. The relative errors of both seen and unseen cases are of the same order of magnitude, which demonstrates the effectiveness of the proposed unsupervised training method. Finally, the streamlines are exhibited in Fig. 12. Compared with the CFD result, the proposed FDGN captures both the central primary and secondary vortices in the lower left and -right corners. Even at lower $Re$, the FDGN method captures the secondary vortex structures successfully. However, the predicted vortex in the lower left corner is smaller than that in the CFD results, which indicates the difficulty of obtaining accurate solutions at relatively high $Re$ (e.g., $Re = 400$).

### 2. Pipe flow around a single cylinder

Solving flows around a circular cylinder is a fundamental case in the literature because of the rich flow patterns that emerge, such as vortex shedding and flow transition. Different from lid-driven cavity flows, the discretization of the physical domain is based on multi-block-structured grids that the CNN-based method cannot handle. Figure 13 shows the block partitioning used for this case.

The domain setting is set the same as the DFG-Benchmark[43] for the single-cylinder case. The domain is $\Omega_{phy} = [0, 2.2] \times [0, 0.41]$. A circular cylinder is considered with center coordinates of $(0.2, 0.2)$ and a diameter of 0.1. The no-slip boundary conditions are still defined on the walls and the inlet velocity $u_{in}$ is defined as a parabolic inflow:

$$u_{in} = \frac{4Uy(0.41 - y)}{0.41^2}, \quad (34)$$

where $U$ defines the maximum velocity and $y$ is the y-axis coordinate. The BCs for the outlet pressure field satisfy Eq. (32).

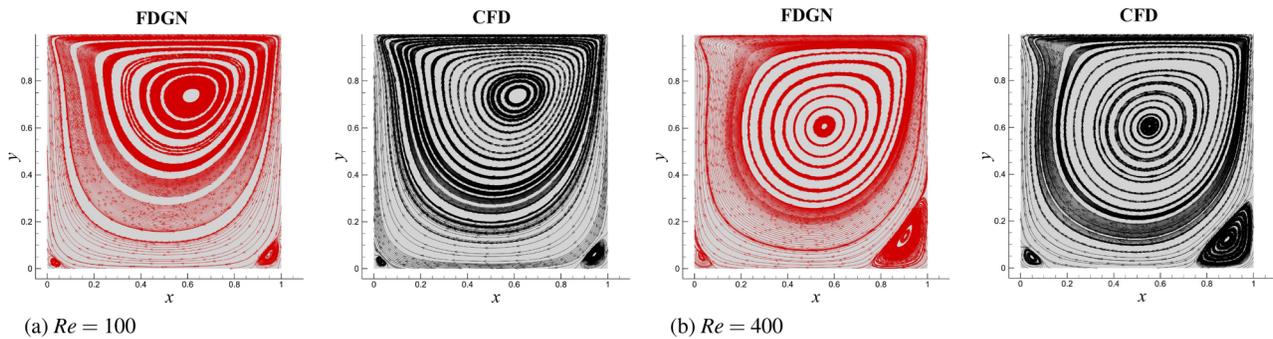

(a) $Re = 100$

(b) $Re = 400$

FIG. 12. Comparison of streamlines for lid-driven cavity flow at different $Re$ between FDGN and CFD. The red and black streamlines denote the results produced by FDGN and CFD, respectively.

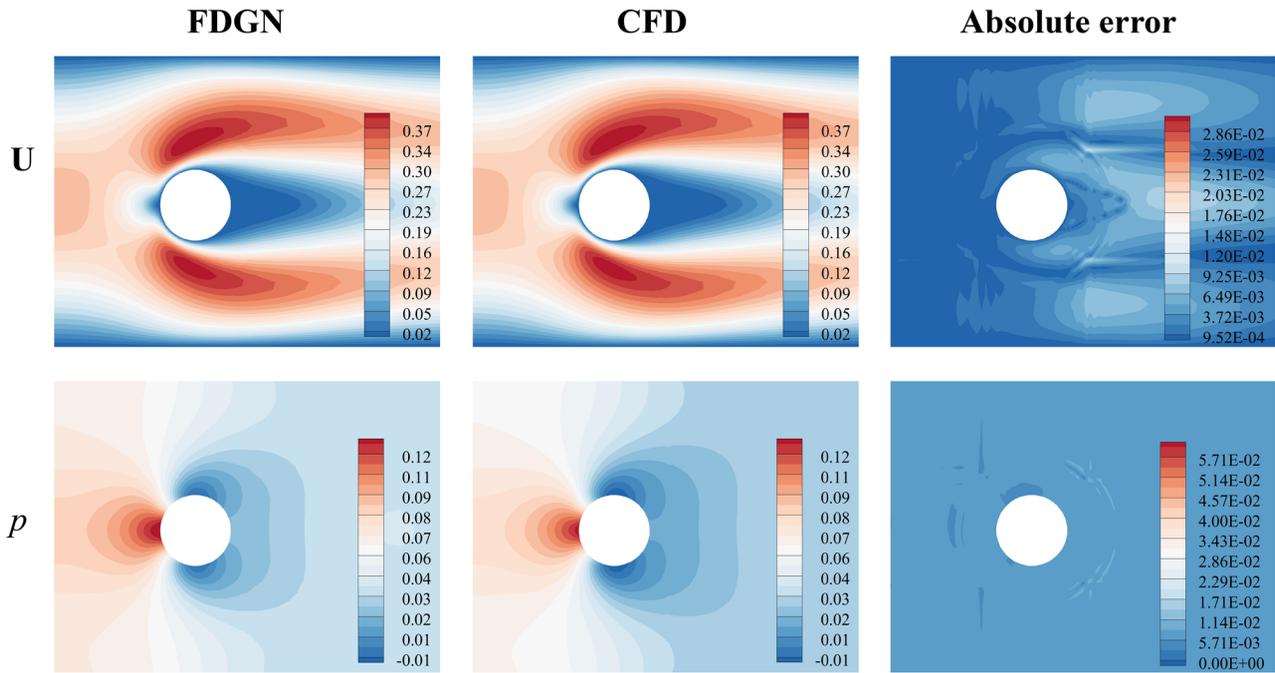

FIG. 14. Contour comparison with the CFD results at $Re = 20$ for pipe flow around a single cylinder.

TABLE III. Drag coefficient ($C_d$) given by different methods. The results for the MAC grid[14] and Spline-PINN[11] are quoted from Ref. 11.

| | $C_d$ at $Re = 20$ | | | | |
|---|---|---|---|---|---|
| Method | MAC grid[14] | Spline-PINN[11] | CFD | DFG-Benchmark[43] | FDGN (Ours) |
| $C_d$ | 4.414 | 4.7 | 5.57 | 5.58 | 5.36 |

The standard single-cylinder case at $Re = 20$ is first compared with the CFD results in Fig. 14. The errors in both the velocity and pressure fields are mainly distributed near the interface between block pairs. This can be attributed to the treatment of the residuals on the interface, as described in Sec. II E 3. The exact interface residuals in physical space are approximated by averaging between blocks that share the same interface, as shown in Fig. 9, which may introduce additional approximation errors. As the proposed GC-FDM is

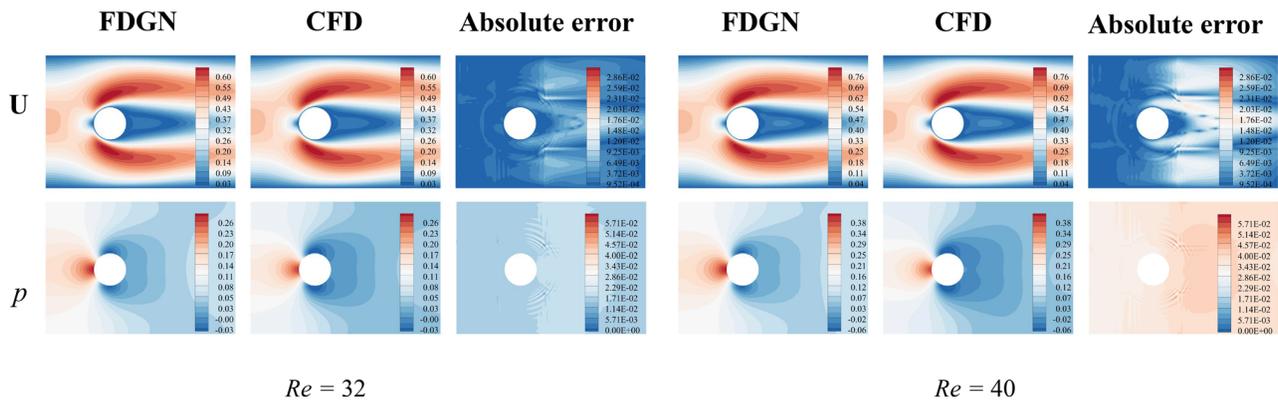

$Re = 32$        $Re = 40$

FIG. 15. Contour comparisons with CFD at the unseen values of $Re = 32$ and $Re = 40$.

conducted in a block-wise manner, it is believed that the resulting errors are strongly related to how the multi-block-structured grid is partitioned. The discontinuity in the grid metrics is most significant between grid blocks around the cylinders, which leads to the approximation errors. This observation demonstrates the difference between FD and FE and indirectly confirms the effectiveness of the proposed FDGN on multi-block-structured grids because it is an FD-based physics-constrained method. In the velocity field, the errors are distributed around the flow separation region, which indicates the difficulty of predicting the flow field under complex meshing and flow patterns. Obtaining solutions on block-structured grids is challenging for image-based methods,[11,14] the superiority of FDGN is now illustrated by quantitatively comparing the drag coefficients ($C_d$) obtained by FDGN at $Re = 20$ with the results using image-based methods,[11,14] CFD, and the official benchmark setup.[43] Table III presents the results. FDGN demonstrates competitive $C_d$ results compared with both CFD and the official benchmark, which confirms the superiority of the proposed method over image-based methods.

The contours for the unseen values of $Re = 32$ and $Re = 40$ are compared in Fig. 15. For this flow case, $Re = 40$ is outside the training range of Refs. 18 and 36, thus allowing us to investigate the model's extrapolation performance. The flow field errors grow sharply from $Re = 32$ to $Re = 40$, both in the interface areas and over the whole domain, which indicates the difficulty of extrapolation.

Table IV depicts the relative flow field errors for this case under both trained and untrained BCs. Similar to the results for the lid-driven cavity flow, the errors in the pressure field are significantly higher than those for the velocity field in each case. FDGN achieves remarkable velocity field predictions, in which the relative error reaches the order of $10^{-3}$, except for the case of $Re = 12$. This may be due to the domination of viscosity at lower flow speeds. In addition, for the unseen cases, the errors in the pressure field are significantly higher than for the trained cases. When extrapolating to $Re = 40$, the relative pressure error exceeds 0.5. This observation verifies the difficulty of extrapolation to BCs outside the training range. As $Re$ approaches 40, the flow changes significantly and some unsteady patterns occur, such as vortex shedding. In this paper, the aim is to solve the steady-state Navier–Stokes equations, which can lead to higher errors.

#### 3. Pipe flow around double cylinders

Now the double-cylinder case is considered to investigate the model performance on a more complex grid. The domain setting is the same as for the single-cylinder case, but with two different cylinders centered at (0.2, 0.16) and (0.35, 0.3) with diameters of 0.09 and 0.06, respectively. Figure 16 shows the block partitioning of this case, which contains more blocks and interfaces than for the single-cylinder configuration.

TABLE IV. Relative mean absolute error for the single-cylinder case at various $Re$.

| Variable | Velocity magnitude | Pressure |
| --- | --- | --- |
| Relative error ($Re = 12$) | 0.023 | 0.39 |
| Relative error ($Re = 20$) | 0.002 | 0.11 |
| Relative error ($Re = 28$) | 0.002 | 0.18 |
| Relative error ($Re = 32$)∗ | 0.003 | 0.50 |
| Relative error ($Re = 36$) | 0.006 | 0.32 |
| Relative error ($Re = 40$)∗ | 0.006 | 0.72 |

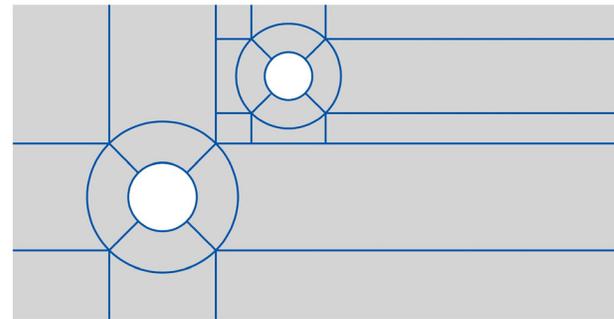

FIG. 16. Block partitioning of the structured grid for the double-cylinder case.

The model predictions at $Re = 30$ are shown in Fig. 17. Similar to the single-cylinder case, the overall domain errors are mainly distributed around the interfaces of the two cylinders. This indicates that the model is more sensitive to these interface regions, where the approximations of grid metrics and Jacobians are not continuous. However, the proposed FDGN is still able to capture the surface distribution of the pressure fields on both cylinders precisely.

To further demonstrate the model performance in trained inlets (BCs) on this more complex gird, the pressure distributions on both cylinders at $Re = 18, 24$, and 30 is presented in Fig. 18. The pressure fields predicted by FDGN fit well with the results from CFD at lower $Re$ (18 and 24). However, as $Re$ increases, the predicted pressure distribution gradually deviates from the CFD results. The interaction between these two cylinders becomes more complex, which may increase the difficulty of predicting the flow field in this double-cylinder case.

To fully illustrate the effectiveness of the proposed method on multi-block-structured grids with complex girds, the inference performance is evaluated for the pipe flow around double cylinders under a series of specific inflow BCs that are not contained in the hybrid dataset. The results are visualized in Fig. 19.

Although untrained at these $Re$ values, FDGN still generalizes well in the extrapolation case and generates faithful flow fields. From the visualization, it is believed that the errors around the discontinuous block areas still demonstrate reasonable generalization ability with these unseen cases using the proposed FD-based method. The relative flow fields errors of these tested cases are presented in Table V. Computational limitations prevent us from training the FDGN sufficiently with richer geometries (varying grid sizes and obstacle positions) and BCs (wider $Re$ ranges and smaller sampling intervals $\delta Re$). Hence, the results verify that the FDGN achieves good inference performance in the FD-based physics-constrained training without vast amounts of labeled data. The relative errors are similar to those for the single-cylinder case, i.e., the pressure field errors are the highest.

In summary, the results from both the single- and double-cylinder cases fully illustrate that the proposed FDGN can effectively handle not only single-block meshes, but more complex multi-block structures. Undoubtedly, obtaining solutions is more challenging in the double-cylinder case.

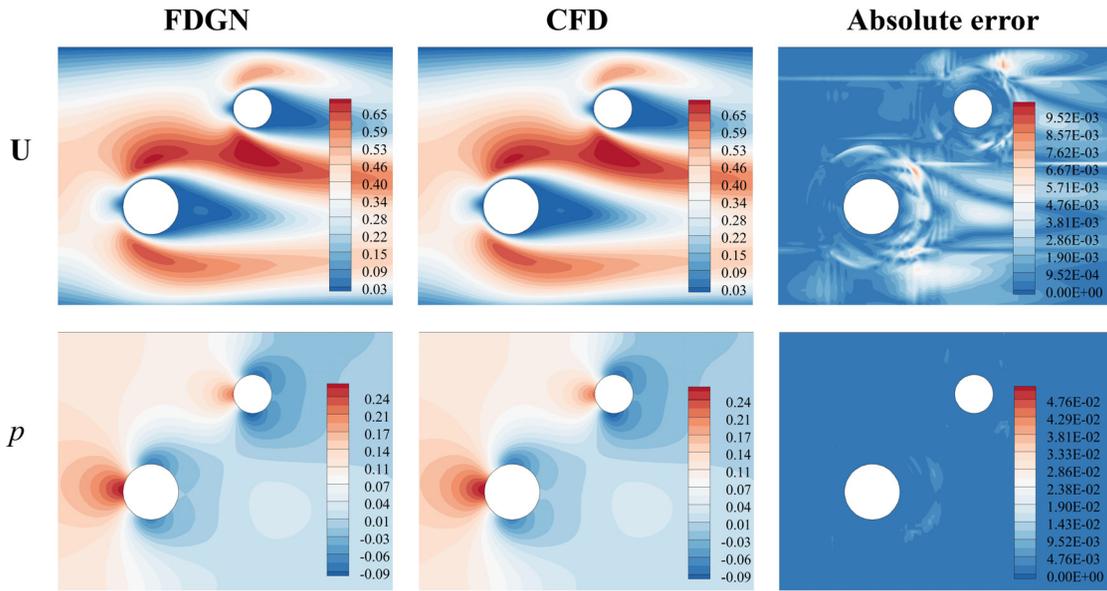

FIG. 17. Contour comparison with CFD at $Re = 30$ for pipe flow around double cylinders.

## B. Comparisons with PINN for nonparameterized solving

In this section, the proposed FDGN with a PINN will be compared in terms of prediction accuracy and training efficiency. As the PINN is more of a uniform framework than a specific technique or methodology, following previous studies[7,28,46] and an FCNN-based PINN is reimplemented, which is the classic design using AD to obtain the PDE residuals. The PINN contains five linear layers with a hidden size of 128. The tanh activation function is employed. The lid-driven flow at $Re = 400$, pipe flow around a single cylinder at $Re = 20$, and pipe flow around double cylinders at $Re = 24$ are chosen as the comparison cases. For consistency, The PINN is fed with coordinates on the same grid. The distribution of collocation points is not investigated in this paper, although this can significantly affect the PINN performance. To achieve a fair comparison with the PINN, FDGN is reimplemented for each case and conduct the training using the corresponding case geometry and fixed BCs. The number of grid samples in the dataset and the minibatch size for training are set to 1 for each case. The model complexity is fixed for both the PINN and FDGN, i.e., the numbers of parameters are 67 000 and 455 000, respectively. Similar to Refs. 7 and 46, the PINN is first trained over several iterations with the Adam optimizer, and then fine-tuned the PINN parameters using the L-BFGS optimization method. The training configuration for each case is described in Table VI.

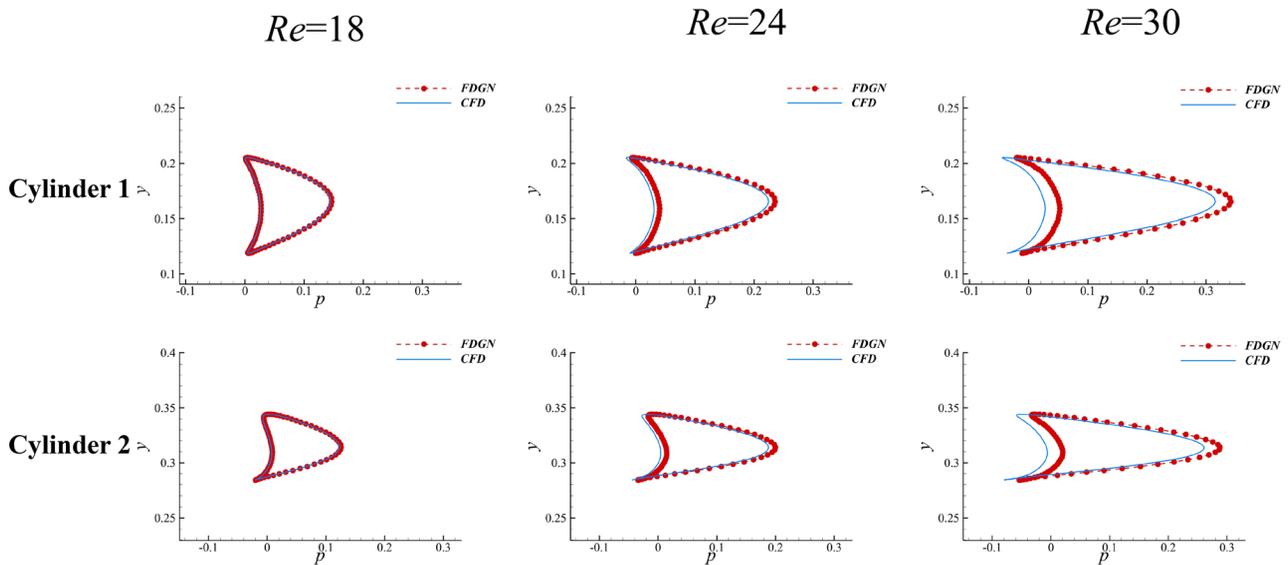

FIG. 18. Comparison of the pressure distributions on both cylinder surfaces at various $Re$. cylinder 1 is the pressure of the larger cylinder at (0.2,0.16) and cylinder 2 is the pressure of the smaller cylinder at (0.35, 0.3).

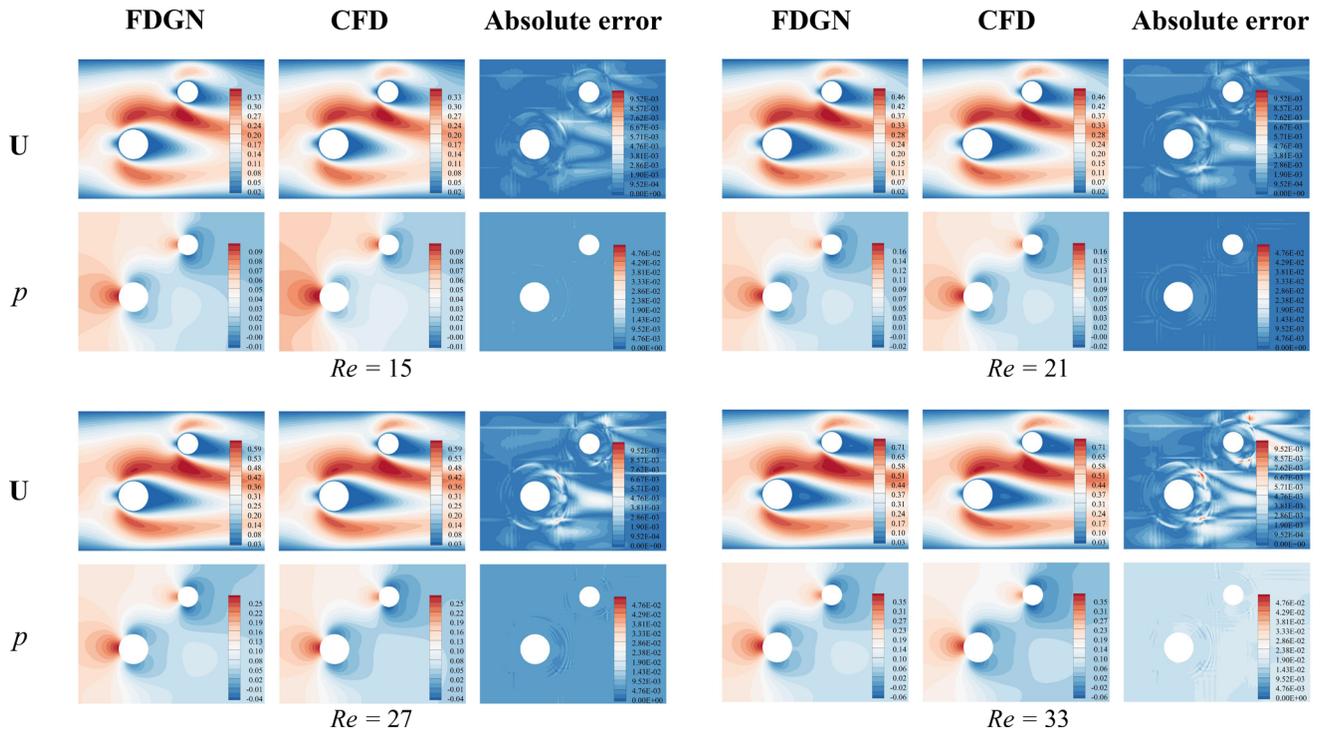

**FIG. 19.** Contour comparisons at various unseen BCs for the double-cylinder case.

**TABLE V.** Relative mean absolute error for the double-cylinder case at various $Re$.

| Variable | Velocity magnitude | Pressure |
| --- | --- | --- |
| Relative error (Re = 15)∗ | 0.005 | 0.35 |
| Relative error (Re = 18) | 0.004 | 0.05 |
| Relative error (Re = 21)∗ | 0.004 | 0.09 |
| Relative error (Re = 24) | 0.004 | 0.13 |
| Relative error (Re = 27)∗ | 0.004 | 0.15 |
| Relative error (Re = 30) | 0.002 | 0.33 |
| Relative error (Re = 33)∗ | 0.005 | 0.37 |

The flow field contours are compared in Fig. 20. For the lid-driven cavity flow, the PINN fails to generate the correct flow fields, so the flow is not able to fully develop. the proposed FDGN shows similar results to the CFD. As discussed in Ref. 26, the performance of PINNs is strongly reliant on the collocation points, and the 3025 points of this grid sample are not sufficient for PINN training. However, our FD-based method is less sensitive to the number of grid points. For the single-cylinder case, the flow fields produced by both FDGN and PINN are in good agreement with the CFD. FDGN achieves similar prediction results to the PINN, which demonstrates the effectiveness of the proposed FD-based physics constraints, although the PINN calculates the derivatives analytically. For the double-cylinder case, although the velocity fields do not reflect the difference between PINN and FDGN, the pressure field produced by FDGN is less accurate than that of the PINN, which demonstrates the advantages of AD in this more complex case compared with the single-cylinder configuration.

**TABLE VI.** Training configurations for PINN and FDGN in each case. These configurations ensure that both models are fully converged for each case. The learning rate (lr) and number of iteration steps ($N_{step}$) are indicated.

| Case | Model | Training (lr, $N_{step}$) | Finetune (lr, $N_{step}$) |
| --- | --- | --- | --- |
| Lid-driven cavity flow at Re = 400. | PINN | Adam: $1 \times 10^{-3}$, $1 \times 10^{5}$ | L-BFGS: N/A, 5 |
| | FDGN | AdamW: $1 \times 10^{-4}$, $2 \times 10^{4}$ | AdamW: $1 \times 10^{-6}$, $1 \times 10^{4}$ |
| Pipe flow around a single cylinder at Re = 20. | PINN | Adam: $1 \times 10^{-3}$, $5 \times 10^{5}$ | L-BFGS: N/A, 10 |
| | FDGN | AdamW: $1 \times 10^{-4}$, $1 \times 10^{4}$ | AdamW: $1 \times 10^{-5}$, $1 \times 10^{4}$ |
| Pipe flow around double cylinders at Re = 24. | PINN | Adam: $1 \times 10^{-3}$, $5 \times 10^{5}$ | L-BFGS: N/A, 10 |
| | FDGN | AdamW: $1 \times 10^{-4}$, $5 \times 10^{4}$ | AdamW: $1 \times 10^{-5}$, $1 \times 10^{4}$ |

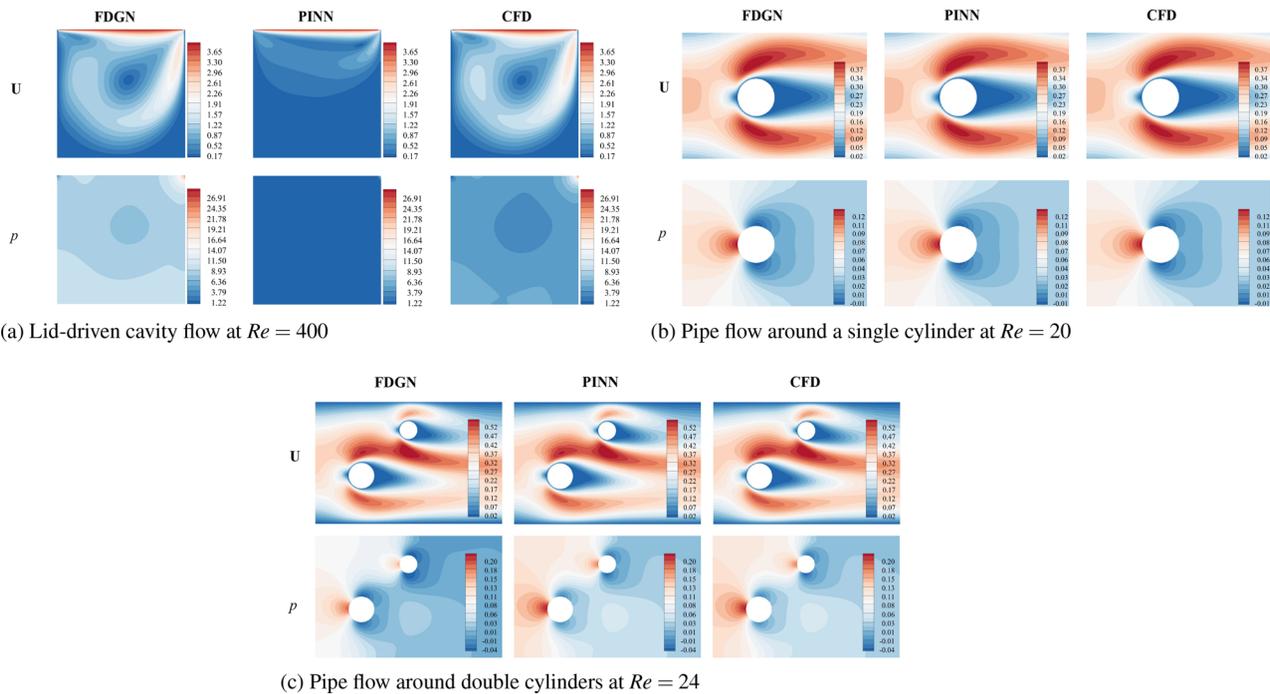

FIG. 20. Flow field comparisons among different methods for the three geometries.

The proposed FD-based method works well with the grid, as the spatial derivatives are approximated in the local stencil. Although the PINN is a meshless method, it may require more collocation points to calculate the derivatives using AD. However, using more grid points always requires more computational resources. Table VII compares the training costs and prediction accuracy against the finite element-based CFD solver. When given sufficient grid points, the PINN achieves remarkable results in both of the pipe flow cases, with a pressure error that is two orders of magnitude lower than that of FDGN in the double-cylinder case. However, the use of AD means that the PINN consumes about 20% more memory than FDGN with the same input and takes more time to converge. As mentioned in Ref. 16, the classic PINN is not suitable for large-scale cases because of the use of AD, which leads to higher computational costs. From the above-mentioned observations, it can be inferred that the FD-based physics constraints make FDGN more resource-efficient, despite having more model parameters, suggesting the potential to be extended to large-scale problems.

### C. Low-speed flow around the 30P30N airfoil

To demonstrate the effectiveness of obtaining solutions on block-structured grids with complex geometries, the performance of FDGN in determining the flow around the 30P30N airfoil geometry is investigated. The physical domain in this case is $\Omega_{phy} = [0, 1.6] \times [0, 0.9]$ for the fully developed far-field flow. The characteristic length of the multi-element high-lift airfoil is fixed to 0.1. Capturing the flow around the airfoil is more challenging than the previous cases considered in this paper because of the highly complex block partitioning with more grid points. The numbers of grid points, blocks, and interfaces are 48 764, 33, and 66, respectively, which is the most complex case considered herein. First, the dramatic increase in the number of grid points increases the training burden for model convergence. The complexity of this geometry entails more curved interfaces and damages the orthogonality of the structured grid, which causes difficulties for the FD approximation in both the Jacobians and PDE residuals. Figure 21 illustrates how the structured grid is partitioned around the 30P30N airfoil, which shows a magnified version of the entire domain.

TABLE VII. Comparison of training costs and prediction accuracy with PINN for each case.

| Case | Model | Velocity error | Pressure error | Training memory consumption | Convergence time |
|---|---|---|---|---|---|
| Lid-driven cavity flow at Re = 400. | PINN | N/A | N/A | 0.6G | N/A |
|  | FDGN | 0.044 | 0.54 | 0.5G | 2.5 h |
| Pipe flow around a single cylinder at Re = 20. | PINN | 0.001 | 0.011 | 1.0G | 4.2 h |
|  | FDGN | 0.003 | 0.007 | 0.8G | 3.5 h |
| Pipe flow around double cylinders at Re = 24. | PINN | 0.001 | 0.004 | 1.6G | 5.6 h |
|  | FDGN | 0.003 | 0.12 | 1.3G | 4.8 h. |

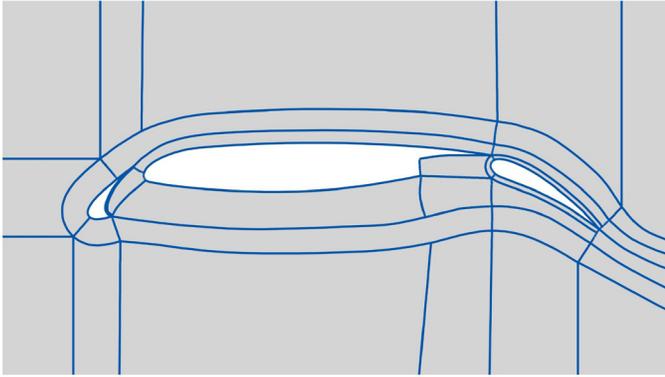

FIG. 21. Block partitioning of the structured grid around the 30P30N airfoil.

Computational limitations mean that parameterized BCs are not solved for this case. Instead, this case is trained and solved with a fixed inflow boundary condition at $Re = 1000$ in a nonparameterized setting. The flow is governed by the uniform inlet velocity $u$ at the left-hand boundary of the rectangular domain. The top, bottom, and right-hand boundaries are assigned outflow BCs. Figure 22 compares the flow fields given by FDGN and CFD. The overall flow structure solved by FDGN agrees well with the results of CFD, although there are some distinct deviations around the flap and wake areas. Although this is a low-speed case, some unsteady flow patterns exist in the wake. In this paper, we only focus on the steady-state incompressible equations, as described in Eq. (10); thus, FDGN fails to capture these patterns. However, this case still indicates that the multi-block processing of FDGN provides good generalization, even with this complex geometry.

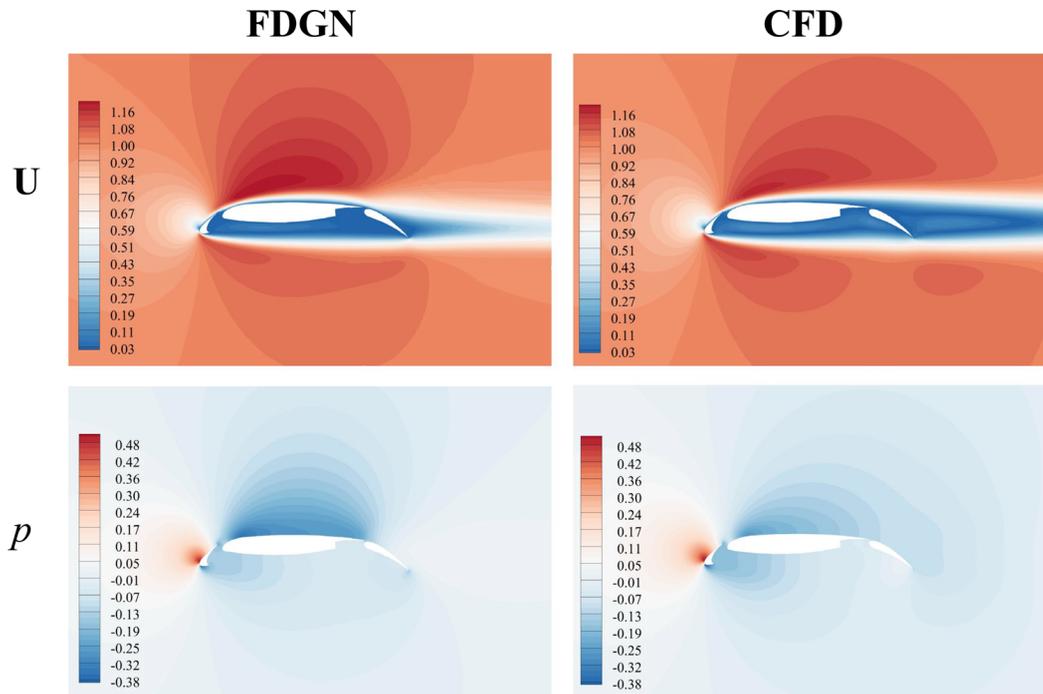

FIG. 22. Contour comparison with CFD at $Re = 1000$ for flow around the 30P30N airfoil.

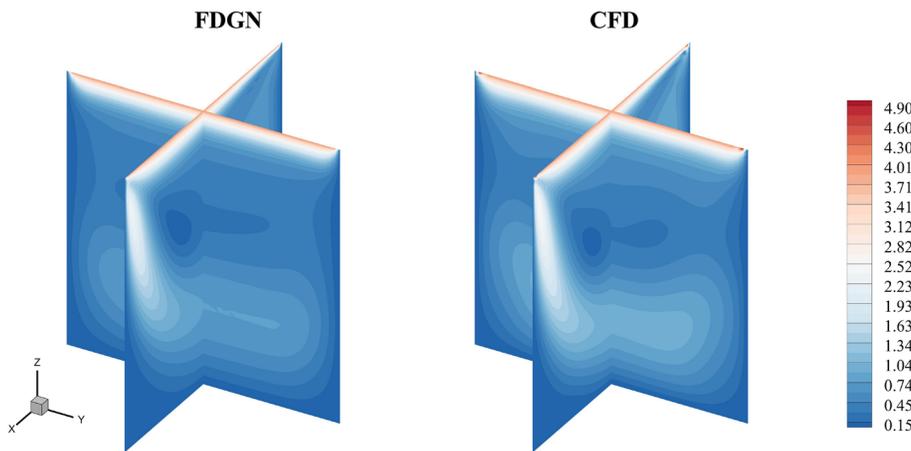

FIG. 23. Comparison of predicted results for the velocity field at $x = 0.5$ and $y = 0.5$ planes.

### D. Applicability in the 3D cavity geometry

All the above-mentioned experiments are conducted in 2D cases. In this subsection, the 3D applicability is simply investigated using the 3D cavity geometry. The 3D physical domain is $\Omega_{phy} = [0,1] \times [0,1] \times [0,1]$ and the grid size is $55 \times 55 \times 55$. After obtaining the node adjacency, the 3D cavity is represented as the graph, and input into the GN. With minor modifications to both the GN model and GC-FDM, the proposed method can be effectively applied to this 3D lid-driven cavity case.

To achieve the validation, the model is trained and evaluated at Reynolds number of 400 in a nonparameterized setting. Figure 23 shows the 3D sectional views of the velocity magnitude $U$ at $x = 0.5$ and $y = 0.5$. As shown in the figure, the predicted velocity field contour closely matches those obtained from the CFD solver, which verifies the model's applicability in this 3D case. To further investigate the validity of the predicted flow field structure, the iso-surface for velocity magnitudes equal to 0.6 is shown in Fig. 24. The visualization of the iso-surface in Fig. 24 indicates a significant deviation between the results of FDGN and the CFD solver. The relative mean absolute error for the overall velocity field is 0.158. In comparison to the 2D case at the same Reynolds number, where the error was 0.037 (as shown in Table II), this represents a considerable increase. This result and the above-mentioned observations highlight the challenges associated with 3D flow prediction. Addressing the 3D problems effectively will require both more suitable numerical scheme and neural network design. Additionally, the increasing grid nodes in 3D not only imposes greater demands on computational resources but also complicates the convergence of the network model.

### IV. DISCUSSION AND CONCLUSION

In this study, an unsupervised GN-based flow prediction method named FDGN is proposed. The aim of this method is to predict flows on block-structured grids, where each block represents an independent curvilinear coordinate system. The main contribution of this work lies in the algorithmic development, whereby FD-based physics constraints are extended to block-structured grids using GN, which enables the solution of parameterized flow fields with complex geometries.

The proposed method demonstrates comparable accuracy and efficiency to existing methods in solving fluid dynamics problems. The FDGN framework effectively combines GNs with the FD method, leveraging the strengths of both approaches to handle complex geometries and various boundary conditions. Through extensive experiments on lid-driven cavity flows and pipe flows around single and double cylinders, FDGN can accurately capture flow patterns and provide reliable predictions, even for unseen BCs. Compared with classic PINNs, FDGN produces significant improvements in computational efficiency and scalability. The FD-based physics constraint enables the model to achieve superior performance with less computational overhead, making it a viable option for large-scale fluid dynamics simulations. Furthermore, experiments reveal that FDGN can handle high-Reynolds number flows with the complex 30P30N geometry, maintaining reasonable velocity and pressure field predictions. This capability is particularly valuable for engineering applications where flow interactions and boundary conditions are highly variable.

Although the proposed FDGN achieves reliable prediction performance, as demonstrated in the above-mentioned results, there are still some limitations of the proposed method. First, there is a large margin for improvement in the accuracy of pressure predictions. Second, errors around interface areas are inevitable in the proposed method due to the discontinuity between blocks, which leads to performance degradation in the FD approximations. Third, the proposed method is limited to steady-state equations, higher Reynolds numbers with suitable numerical time-stepping schemes as well as turbulent models would be worth exploring in the future work. Finally, the proposed GC-FDM is limited in 2D multi-block-structured grids and it has the potential to be extended to more complex 3D geometries.

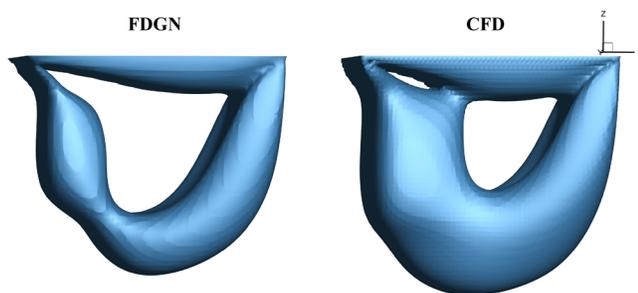

FIG. 24. Comparison of predicted results for the velocity iso-surface at $U = 0.6$.

In summary, the FDGN framework represents a significant advance in the intersection of computational fluid dynamics and neural networks. By integrating GN and FD methods, it offers a robust and scalable solution for predicting fluid flows on block-structured grids. Future work will focus on further optimizing the model for larger and more complex simulations, as well as exploring its application to other types of PDEs beyond fluid dynamics.

## ACKNOWLEDGMENTS

This research is funded by the National Key Project of China (Grant No. GJXM92579) and is also supported by the Sichuan Science and Technology Program (No. 2023YFG0329 and Project No. 2021ZDZX0001).

## AUTHOR DECLARATIONS
### Conflict of Interest

The authors have no conflicts to disclose.

### Author Contributions

**Yiye Zou:** Conceptualization (equal); Data curation (equal); Formal analysis (equal); Investigation (equal); Methodology (equal); Software (equal); Visualization (equal); Writing – original draft (equal). **Tianyu Li:** Conceptualization (equal); Formal analysis (equal); Investigation (equal); Software (equal); Validation (equal); Visualization (equal). **Lin Lu:** Formal analysis (equal); Software (equal); Validation (equal). **Jingyu Wang:** Project administration (equal); Resources (equal); Supervision (equal); Writing – review & editing (equal). **Shufan Zou:** Supervision (equal). **Laiping Zhang:** Resources (equal); Supervision (equal). **Xiaogang Deng:** Project administration (lead); Resources (lead); Supervision (equal).

## DATA AVAILABILITY

The data that support the findings of this study are available from the corresponding author upon reasonable request.

## REFERENCES

[1] F. d A. Belbute-Peres, T. D. Economon, and J. Z. Kolter, "Combining differentiable PDE solvers and graph neural networks for fluid flow prediction," in International Conference on Machine Learning (ICML), 2020.
[2] J. Brandstetter, D. Worrall, and M. Welling, "Message passing neural PDE solvers," arXiv:2202.03376 (2022).
[3] L.-W. Chen and N. Thuerey, "Towards high-accuracy deep learning inference of compressible flows over aerofoils," Comput. Fluids 250, 105707 (2023).
[4] Z. Li, N. Kovachki, K. Azizzadenesheli, B. Liu, K. Bhattacharya, A. Stuart, and A. Anandkumar, "Fourier neural operator for parametric partial differential equations," arXiv:2010.08895 (2020).
[5] L. Lu, P. Jin, G. Pang, Z. Zhang, and G. E. Karniadakis, "Learning nonlinear operators via DeepONet based on the universal approximation theorem of operators," Nat. Mach. Intell. 3, 218–229 (2021).
[6] T. Pfaff, M. Fortunato, A. Sanchez-Gonzalez, and P. W. Battaglia, "Learning mesh-based simulation with graph networks," in International Conference on Learning Representations, 2021.
[7] M. Raissi, P. Perdikaris, and G. E. Karniadakis, "Physics-informed neural networks: A deep learning framework for solving forward and inverse problems involving nonlinear partial differential equations," J. Comput. Phys. 378, 686–707 (2019).
[8] C. Rao, H. Sun, and Y. Liu, "Physics-informed deep learning for incompressible laminar flows," Theor. Appl. Mech. Lett. 10, 207–212 (2020).
[9] X. Jin, S. Cai, H. Li, and G. E. Karniadakis, "NSFnets (Navier–Stokes flow nets): Physics-informed neural networks for the incompressible Navier–Stokes equations," J. Comput. Phys. 426, 109951 (2021).
[10] L. Sun, H. Gao, S. Pan, and J.-X. Wang, "Surrogate modeling for fluid flows based on physics-constrained deep learning without simulation data," Comput. Methods Appl. Mech. Eng. 361, 112732 (2020).
[11] N. Wandel, M. Weinmann, M. Neidlin, and R. Klein, "Spline-PINN: Approaching PDEs without data using fast, physics-informed Hermite-Spline CNNs," in Proceedings of the AAAI Conference on Artificial Intelligence (AAAI, 2022), Vol. 36, pp. 8529–8538.
[12] A. G. Baydin, B. A. Pearlmutter, A. A. Radul, and J. M. Siskind, "Automatic differentiation in machine learning: A survey," J. Mach. Learn. Res. 18, 1–43 (2018).
[13] A. Paszke, S. Gross, S. Chintala, G. Chanan, E. Yang, Z. DeVito, Z. Lin, A. Desmaison, L. Antiga, and A. Lerer, "Automatic differentiation in PyTorch," 2017.
[14] N. Wandel, M. Weinmann, and R. Klein, "Learning incompressible fluid dynamics from scratch—Towards fast, differentiable fluid models that generalize," 2021.
[15] N. Wandel, M. Weinmann, and R. Klein, "Teaching the incompressible Navier–Stokes equations to fast neural surrogate models in 3D," Phys. Fluids 33, 047117 (2021).
[16] H. Gao, L. Sun, and J.-X. Wang, "PhyGeoNet: Physics-informed geometry-adaptive convolutional neural networks for solving parameterized steady-state PDEs on irregular domain," J. Comput. Phys. 428, 110079 (2021).
[17] Y. Zang, R. L. Street, and J. R. Koseff, "A non-staggered grid, fractional step method for time-dependent incompressible Navier–Stokes equations in curvilinear coordinates," J. Comput. Phys. 114, 18–33 (1994).
[18] C. Hirsch, Numerical Computation of Internal and External Flows: The Fundamentals of Computational Fluid Dynamics (Elsevier, 2007).
[19] J. H. Ferziger, M. Perić, and R. L. Street, Computational Methods for Fluid Dynamics (Springer, 2019).
[20] S. Mazumder, Numerical Methods for Partial Differential Equations: Finite Difference and Finite Volume Methods (Academic Press, 2015).
[21] P. W. Battaglia, J. B. Hamrick, V. Bapst, A. Sanchez-Gonzalez, V. Zambaldi, M. Malinowski, A. Tacchetti, D. Raposo, A. Santoro, R. Faulkner et al., "Relational inductive biases, deep learning, and graph networks," arXiv:1806.01261 (2018).
[22] M. Fortunato, T. Pfaff, P. Wirnsberger, A. Pritzel, and P. Battaglia, "Multiscale meshgraphnets," arXiv:2210.00612 (2022).
[23] X. Han, H. Gao, T. Pfaff, J.-X. Wang, and L.-P. Liu, "Predicting physics in mesh-reduced space with temporal attention," arXiv:2201.09113 (2022).
[24] A. Sanchez-Gonzalez, J. Godwin, T. Pfaff, R. Ying, J. Leskovec, and P. W. Battaglia, "Learning to simulate complex physics with graph networks," in International Conference on Machine Learning, 2020.
[25] D. Chen, Y. Lin, W. Li, P. Li, J. Zhou, and X. Sun, "Measuring and relieving the over-smoothing problem for graph neural networks from the topological view," in Proceedings of the AAAI Conference on Artificial Intelligence (AAAI, 2020), Vol. 34, pp. 3438–3445.
[26] P.-H. Chiu, J. C. Wong, C. Ooi, M. H. Dao, and Y.-S. Ong, "CAN-PINN: A fast physics-informed neural network based on coupled-automatic–numerical differentiation method," Comput. Methods Appl. Mech. Eng. 395, 114909 (2022).
[27] S. Tang, X. Feng, W. Wu, and H. Xu, "Physics-informed neural networks combined with polynomial interpolation to solve nonlinear partial differential equations," Comput. Math. Appl. 132, 48–62 (2023).
[28] W. Cao, J. Song, and W. Zhang, "A solver for subsonic flow around airfoils based on physics-informed neural networks and mesh transformation," Phys. Fluids 36, 027134 (2024).
[29] O. Ronneberger, P. Fischer, and T. Brox, "U-net: Convolutional networks for biomedical image segmentation," in Medical Image Computing and Computer-Assisted Intervention–MICCAI 2015: 18th International Conference, Munich, Germany, 5–9 October, Proceedings, Part III 18 (Springer, 2015), pp. 234–241.
[30] K. L. Lim, R. Dutta, and M. Rotaru, "Physics informed neural network using finite difference method," in 2022 IEEE International Conference on Systems, Man, and Cybernetics (SMC) (IEEE, 2022), pp. 1828–1833.

[31] D. Xu, X. Deng, Y. Chen, G. Wang, and Y. Dong, "Effect of nonuniform grids on high-order finite difference method," Adv. Appl. Math. Mech. 9, 1012–1034 (2017).

[32] X. Deng, M. Mao, G. Tu, H. Liu, and H. Zhang, "Geometric conservation law and applications to high-order finite difference schemes with stationary grids," J. Comput. Phys. 230, 1100–1115 (2011).

[33] X. Deng, Y. Min, M. Mao, H. Liu, G. Tu, and H. Zhang, "Further studies on geometric conservation law and applications to high-order finite difference schemes with stationary grids," J. Comput. Phys. 239, 90–111 (2013).

[34] T. Li, S. Zou, X. Chang, L. Zhang, and X. Deng, "Predicting unsteady incompressible fluid dynamics with finite volume informed neural network," Phys. Fluids 36, 043601 (2024a).

[35] T. Li, Y. Zou, S. Zou, X. Chang, L. Zhang, and X. Deng, "A fully differentiable GNN-based PDE solver: With applications to Poisson and Navier–Stokes equations," arXiv:2405.04466 (2024).

[36] S. Elfwing, E. Uchibe, and K. Doya, "Sigmoid-weighted linear units for neural network function approximation in reinforcement learning," Neural Networks 107, 3–11 (2018).

[37] J. Gilmer, S. S. Schoenholz, P. F. Riley, O. Vinyals, and G. E. Dahl, "Neural message passing for quantum chemistry," in International Conference on Machine Learning (PMLR, 2017), pp. 1263–1272.

[38] W. Hamilton, Z. Ying, and J. Leskovec, "Inductive representation learning on large graphs," in Advances in Neural Information Processing Systems (Curran Associates, Inc., 2017), Vol. 30.

[39] S. Brody, U. Alon, and E. Yahav, "How attentive are graph attention networks?," in International Conference on Learning Representations, 2022.

[40] T. N. Kipf and M. Welling, "Semi-supervised classification with graph convolutional networks," arXiv:1609.02907 (2016).

[41] P. Velickovic, G. Cucurull, A. Casanova, A. Romero, P. Lio, Y. Bengio et al., "Graph attention networks," stat 1050, 10–48550 (2017).

[42] M. Fey and J. E. Lenssen, "Fast graph representation learning with PyTorch geometric," in ICLR Workshop on Representation Learning on Graphs and Manifolds, 2019.

[43] TU Dortmund University, "DFG benchmarking on laminar flow around a cylinder," https://www.mathematik.tu-dortmund.de/~featflow/en/benchmarks/cfdbenchmarking/flow/dfg_benchmark1_re20.html; accessed 21 September 2024.

[44] I. Loshchilov and F. Hutter, "Decoupled weight decay regularization," arXiv:1711.05101 (2017).

[45] A. Paszke, S. Gross, F. Massa, A. Lerer, J. Bradbury, G. Chanan, T. Killeen, Z. Lin, N. Gimelshein, L. Antiga et al., "PyTorch: An imperative style, high-performance deep learning library," in Advances in Neural Information Processing Systems (Curran Associates, Inc., 2019), Vol. 32.

[46] S. Wang, S. Sankaran, H. Wang, and P. Perdikaris, "An expert's guide to training physics-informed neural networks," arXiv:2308.08468 (2023).